\documentclass[10pt,twocolumn,letterpaper]{article}
\usepackage[pagenumbers]{cvpr}
\usepackage[T1]{fontenc}
\usepackage{amsthm}
\usepackage{multirow}
\usepackage{colortbl}
\definecolor{rebatint}{HTML}{F0F6EC}
\usepackage{algorithm}
\usepackage{algorithmic}
\usepackage{placeins}
\usepackage{xr}
\definecolor{cvprblue}{rgb}{0.21,0.49,0.74}
\usepackage[pagebackref,breaklinks,colorlinks,allcolors=cvprblue]{hyperref}
\def\confName{CVPR}
\def\confYear{2027}

\newcommand{\stdaux}{\textsc{Std-Aux}}
\newcommand{\stdauxsub}{\mathrm{Std\text{-}Aux}}
\newcommand{\Laux}{\mathcal{L}_{\mathrm{aux}}}

\newcommand{\Linst}{\mathcal{L}_{\mathrm{inst}}}

\newcommand{\method}{ReBA}
\newcommand{\methodfull}{Relax Within, Balance Across}
\newcommand{\Lreba}{\mathcal{L}_{\mathrm{ReBA}}}

\newcommand{\cmark}{\ensuremath{\checkmark}}
\newcommand{\xmark}{\ensuremath{\times}}

\newcommand{\ReBAArxiv}{}
\title{\methodfull: Load Balancing under Composition Shifts\\
in Vision-Language Mixture-of-Experts}
\author{
  Ziang Wu\textsuperscript{2,3}\thanks{These authors contributed equally.}\quad
  Peng Jin\textsuperscript{1,4,*}\quad
  Qishen Yin\textsuperscript{1,*}\quad
  Munan Ning\textsuperscript{1,2}\\[3pt]
  Hao Li\textsuperscript{1,4}\quad
  Peizhen Zhang\textsuperscript{5}\quad
  Huihui Zhou\textsuperscript{2}\quad
  Li Yuan\textsuperscript{1,2}\thanks{Corresponding author.}\\[6pt]
  {\small\textsuperscript{1}Shenzhen Graduate School, Peking University}\\
  {\small\textsuperscript{2}Peng Cheng Laboratory}\\
  {\small\textsuperscript{3}School of Software and Microelectronics, Peking University}\\
  {\small\textsuperscript{4}Qwen Team\qquad\textsuperscript{5}Sun Yat-sen University}\\[4pt]
  {\footnotesize\{ziangwu, giesen\_yin\}@stu.pku.edu.cn,\quad
    \{jp21mails, munanning, ozymandisli\}@gmail.com,}\\
  {\footnotesize zhangpzh5@alumni.sysu.edu.cn,\quad
    zhouhh@pcl.ac.cn,\quad yuanli-ece@pku.edu.cn}
}

\hypersetup{
  pdftitle={Relax Within, Balance Across: Load Balancing under Composition Shifts in Vision-Language Mixture-of-Experts},
  pdfauthor={Ziang Wu, Peng Jin, Qishen Yin, Munan Ning, Hao Li, Peizhen Zhang, Huihui Zhou, Li Yuan}
}
\begin{document}
\maketitle
\begin{figure*}[!t]
\centering
\includegraphics[width=\textwidth]{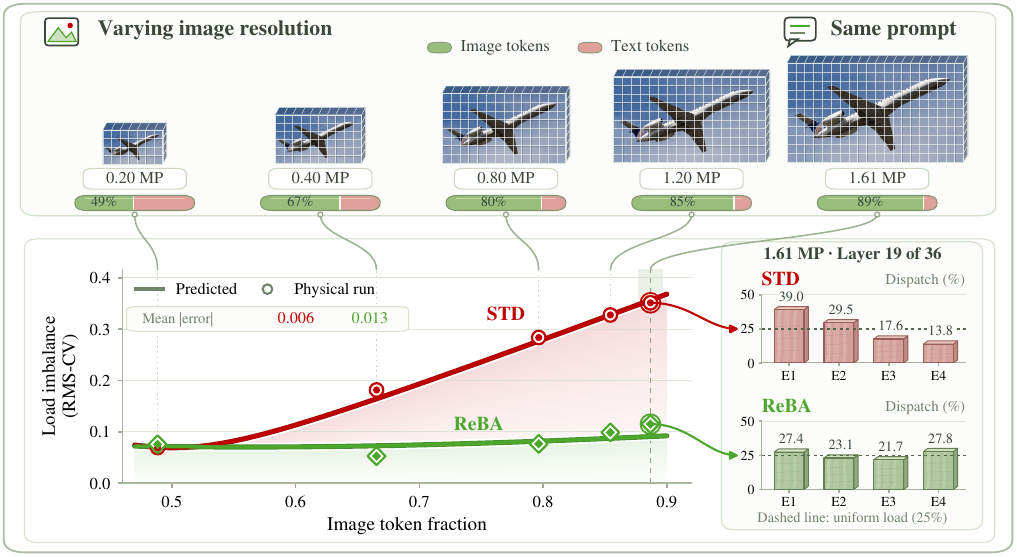}
\caption{Predicting how expert load changes with image resolution.
Split-Qwen3VL-4B processes the same 466 requests at five resolutions.
Capsules show image and text token proportions; percentages give the image fraction.
Curves use modality routing profiles from one reference condition.
Markers show physical runs, and short links show prediction errors.
Mean $|\mathrm{error}|$ averages absolute RMS-CV errors over the five runs.
ReBA keeps load more stable as image tokens increase.
Right: layer-19 dispatch shares at the highlighted resolution.
The photo and grids illustrate resolution changes. Lower RMS-CV is better.}
\label{fig:teaser}
\end{figure*}

\begin{abstract}

Vision-language mixture-of-experts (MoE) models such as Qwen3-VL need balanced
loads for fast parallel serving. Standard training pools image and text tokens
into one auxiliary loss, assuming training balance holds at inference. Yet we
find a surprising mismatch: on a fixed router, balance shifts sharply whenever
resolution, dynamic tiling, or image count changes the image-to-text token
ratio. Splitting load by modality reveals \emph{cross-modal load cancellation}:
opposing image and text errors cancel near the training ratio, hiding strong
skew until the mix changes. To quantify this shift, we derive an exact quadratic
law: under fixed modality profiles, one reference run predicts the full response
curve, and the image--text gap governs sensitivity. Meanwhile, patches within an
image route similarly while patch counts vary, letting large images dominate
flat pooling. We propose \method{} (\methodfull) to separate modality losses by
token fraction and weight images equally, letting each image specialize while
keeping overall visual load even. Across four split backbones, \method{} turns
fragile single-ratio balance into stable balance across resolutions, tilings,
and image counts without hurting accuracy. Large modality gaps in released
native MoEs also show its promise for native post-training.
\urlstyle{same}
Code is available at \url{https://github.com/ZiangWu-77/ReBA}.
\end{abstract}

\section{Introduction}
\label{sec:intro}

Vision-language mixture-of-experts (MoE) models route visual patches and text
tokens to sparse experts~\cite{riquelme2021vmoe,lin2024moellava,wu2024deepseekvl2}
and rely on auxiliary losses to balance expert
load~\cite{shazeer2017outrageously,fedus2022switch}. Standard training pools both modalities into one loss, assuming training balance
holds at inference. Yet in deployment, we find a sharp train--inference
mismatch: expert balance shifts markedly with the image-to-text token mix.
Because resolution, dynamic tiling, and image count directly alter this
mix~\cite{wang2024qwen2vl,chen2024internvl15}, routine serving changes destabilize
load, echoing hotspots noted in serving systems~\cite{li2026macs}: in
Fig.~\ref{fig:teaser}, varying resolution alone on Split-Qwen3VL-4B shifts
imbalance by over fivefold. Why does the same fixed router change its load when
only the token mix changes?

\begin{figure*}[t]
\centering
\includegraphics[width=0.98\textwidth]{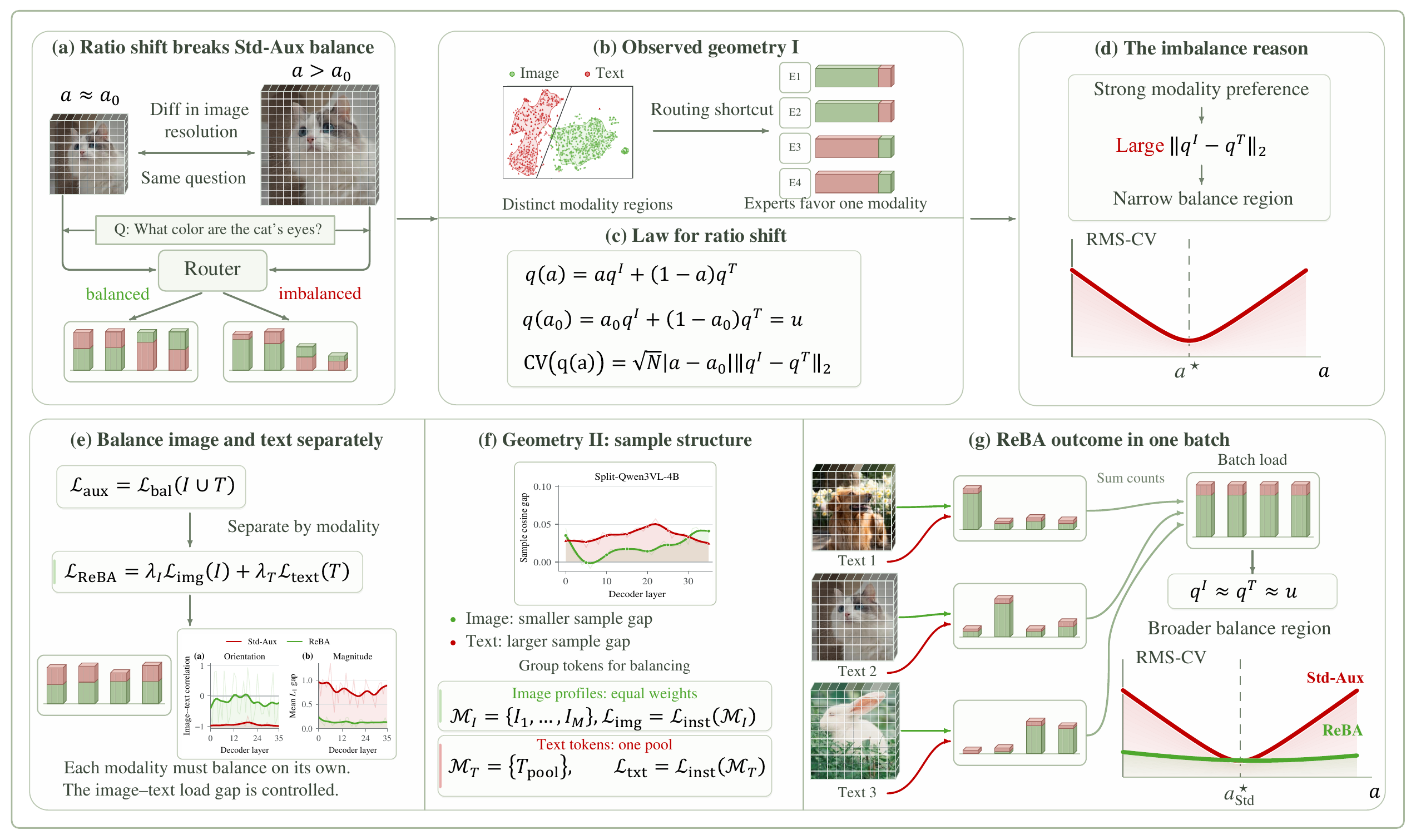}
\caption{From routing structure to ReBA.
(a--d) Image and text loads can cancel at one token mix. Conditional profiles
from one reference condition predict how load changes with that mix.
(e) \method{} balances image and text separately, reducing their load gap.
(f) Sample structure motivates one equally weighted profile per source image and
one pooled text profile; the visual loss is computed after averaging the image
profiles.
(g) Schematic batch dispatches illustrate distinct image preferences that offset
across a batch. Response curves in (d,g) use the reference profiles from
Fig.~\ref{fig:teaser}. In (e,f), faint curves show layer values and solid lines show trends; the
line in (b) is a visual guide.}
\label{fig:method}
\end{figure*}

To explain why training balance fails at inference, we profile expert dispatch
per modality and uncover \emph{cross-modal load cancellation}
(Fig.~\ref{fig:method}(a--c)). Under a pooled loss, image and text tokens
develop opposing preferences that offset near the training mix (Appendix~\ref{sup:training-composition}), masking strong
per-modality skew until the inference mix shifts. Further, we find that this
drift is not erratic but follows an exact law: under fixed conditional profiles,
a single reference condition predicts the full load-response curve across mixes
(Fig.~\ref{fig:teaser}, Fig.~\ref{fig:method}(d)). We derive that squared aggregate imbalance is an exact quadratic in the
image-token fraction, with curvature governed by the squared distance between
reference modality profiles. Apparent balance at one mix thus coexists with high
sensitivity when this gap is large.

Stabilizing load across token mixes therefore requires shrinking the image--text
load gap during training. Yet inference-time schedulers such as
MACS~\cite{li2026macs} only adapt runtime capacity around fixed hotspots rather
than correcting the router, whereas training-time modality regularizers either
encourage entropy against minority-modality collapse
(LIMoE~\cite{mustafa2022limoe}) or bound routing divergence in instruction
tuning to retain language skill (SMAR~\cite{xia2025smar}). Because neither
aligns realized image and text load profiles, the router can still retain a
large cross-modal gap.

During supervised fine-tuning, \method{} (\methodfull) shrinks this gap through
two design choices grounded in Fig.~\ref{fig:method}(e--g). First, it weights
separate image and text balance losses by token fractions so opposing modality
errors cannot cancel (Fig.~\ref{fig:method}(e)). Second, because patches within
an image are more similar in router space than across images
(Fig.~\ref{fig:method}(f)) and patch counts vary, flat pooling lets large images
dominate the visual loss. \method{} instead averages routing profiles per image
and weights images equally in one visual loss (Fig.~\ref{fig:method}(f,g)),
balancing the mean image profile while letting individual images favor different
experts. Valid text tokens in the device-local microbatch are pooled equally.

Across four split vision-language backbones
\cite{zhang2022moefication,zhu2024llamamoe}, \method{} lowers benchmark load
imbalance with comparable average task accuracy. On Split-Qwen3VL-4B, reference
profiles closely predict the physical resolution sweep (Fig.~\ref{fig:teaser}),
and \method{} lowers imbalance across resolutions and across image counts, where
it also improves over adapted LIMoE-style modality entropy
\cite{mustafa2022limoe} on the four-expert model. \method{} likewise lowers
imbalance across dynamic tiling on Split-InternVL3-8B. Diagnostics on two
released native MoEs also show large image--text load gaps
(Appendix~\ref{sup:native}), which suggests that \method{} may improve balance
during native multimodal MoE training.

\begin{samepage}
Our main contributions are:
\begin{itemize}
\item We uncover \emph{cross-modal load cancellation} under pooled balancing,
explaining why token-mix shifts destabilize fixed routers, and diagnose large
image--text load gaps in native MoEs.
\item We derive an exact quadratic law showing that one reference condition
predicts the full load-response curve from the image--text profile gap, verified
on physical resolution sweeps.
\item We propose \method{}, weighting separate modality losses by token
fractions and image profiles equally while pooling text tokens, lowering
imbalance across mixes at comparable accuracy on four split backbones.
\end{itemize}
\end{samepage}
\section{Background and Related Work}
\label{sec:background}

\paragraph{Sparse VLMs and pooled balancing.}
A sparse MoE layer routes each token $x_t$ to $k$ of $N$ experts using router
logits $g(x_t)$ \cite{shazeer2017outrageously,lepikhin2021gshard,fedus2022switch}.
For expert $i$, $f_{t,i}=k^{-1}\mathbf{1}[i\in\mathrm{top\text{-}}k(g(x_t))]$
records hard dispatch and $p_{t,i}=\mathrm{softmax}(g(x_t))_i$ gives the soft
gate probability. Averaging over $T$ valid tokens yields $f_i=T^{-1}\sum_t f_{t,i}$
and $p_i=T^{-1}\sum_t p_{t,i}$ in the Switch loss \cite{fedus2022switch}:
\begin{equation}
\Laux=N\sum_{i=1}^{N} f_i p_i.
\label{eq:laux}
\end{equation}
Here $f_i$ measures realized load and $p_i$ supplies router gradients. Sparse VLMs trained natively \cite{wu2024deepseekvl2}, upcycled
\cite{komatsuzaki2023upcycling,lin2024moellava,li2024cumo,li2024unimoe}, or
split from dense FFN blocks \cite{zhang2022moefication,zhu2024llamamoe} typically
apply Eq.~\eqref{eq:laux} to the pooled token stream, constraining balance only
at the observed token mix.

\paragraph{Modality-aware routing.}
Prior multimodal MoE methods handle modality differences through architecture or
routing regularization. MoMa \cite{lin2024moma} assigns separate expert pools per
modality with expert-choice routing, CuMo \cite{li2024cumo} co-upcycles the
vision encoder and connector, and Uni-MoE \cite{li2024unimoe} uses staged
alignment. In a shared expert pool, LIMoE \cite{mustafa2022limoe} shows that
pooled balancing can hide minority-modality collapse under variable token counts
in contrastive learning, adding local token entropy and thresholded per-modality
global entropy losses to keep a minimum of experts active. For generative VLMs,
SMAR \cite{xia2025smar} bounds the symmetric KL divergence between modality
routing distributions to encourage specialization while retaining language
skills, and Bo et al.\ \cite{bo2026smoes} use soft modality scores and expert
bins to guide specialization. Our work analyzes how the realized image--text
load gap governs sensitivity to token-mix shifts and aligns both profiles
through modality separation and per-image visual aggregation.

\paragraph{Aggregation scope and runtime systems.}
In language MoEs, Qiu et al.\ \cite{qiu2025globalbatch} aggregate dispatch counts
over the global batch to aid specialization, while loss-free balancing
\cite{wang2024auxfreebalancing} updates routing biases from recent loads. Our
fixed-profile law applies to profiles from any training method, while \method{}
groups local microbatch statistics by modality and image. At inference,
expert-parallel serving relies on balanced loads to avoid stragglers
\cite{he2021fastmoe,rajbhandari2022deepspeedmoe,hwang2022tutel,gale2022megablocks,he2026capacityaware}.
For multimodal serving on pretrained Qwen3-VL, InternVL3.5, and Kimi-VL, MACS
\cite{li2026macs} scales capacity from entropy-weighted visual load and modality
mix. ReaLB \cite{wang2026realb} lowers precision on overloaded expert ranks.

\section{Composition-Sensitive Expert Load}
\label{sec:phenomenon}

\subsection{A Fixed Router, a Changing Load}
The resolution sweep in Fig.~\ref{fig:teaser} changes \stdaux{} RMS-CV by more
than fivefold with all model parameters held fixed. We therefore examine image
and text loads separately to explain how changing the token mix shifts overall
imbalance.

\subsection{Modality-Complementary Loads}
\label{sec:phenomenon-shortcut}

Let $q_l^I,q_l^T$ denote the image and text load profiles at MoE layer $l$,
averaging hard dispatch $f_t$ over valid tokens of each modality, with mean
conditional gap $G_1=L^{-1}\sum_l\lVert q_l^I-q_l^T\rVert_1$ across $L$
layers. Profiles are \emph{modality complementary} when their load errors
oppose each other.

On Split-Qwen3VL-4B, \stdaux{} yields a mean image--text Pearson correlation of
$-0.949$ and gap $G_1=0.798$, yet a mixed mean layer CV of $0.273$ as opposing
preferences cancel at the probe ratio. \method{} reduces this opposition,
bringing correlation to $-0.166$ and $G_1$ to $0.137$
(Fig.~\ref{fig:method}(e)). Two released native MoEs in
Appendix~\ref{sup:native} (Qwen3-VL-MoE-30B and Qwen3.5-MoE-35B) likewise show
large conditional gaps, with conditional and mixed RMS-CVs over $23\times$ the
count-matched uniform null mean. Next, we derive how this gap governs the load
response.

\subsection{An Exact Composition-Shift Law}
\label{sec:phenomenon-law}
\label{sec:method-mixture}

We first hold the conditional profiles $q^I,q^T$ fixed to isolate the
image-token fraction $a$. Measured at one reference condition, these two profiles
determine the mixed load $q(a)=a q^I+(1-a)q^T$. Write $u=(1/N,\ldots,1/N)$ for
uniform expert use. If the router is balanced at a reference composition $a_0$,
so $q(a_0)=u$, then
\begin{equation}
\boxed{q(a)-u=(a-a_0)(q^I-q^T).}
\label{eq:ratio-shift}
\end{equation}
The load change thus factors into a composition shift and a conditional gap, so
balance at $a_0$ can coexist with a large gap that amplifies ratio shifts.

The general case also allows a nonzero minimum. With
$\mathrm{CV}(q)=\sqrt{N}\lVert q-u\rVert_2$, define squared aggregate RMS-CV as
$R(a)=L^{-1}\sum_l\mathrm{CV}^2(q_l(a))$, so the plotted RMS-CV curve is
$\sqrt{R(a)}$. Expanding and completing the square gives
\begin{equation}
\boxed{\begin{aligned}
R(a)&=R(a^\star)+\kappa(a-a^\star)^2,\\
\kappa&=\frac NL\sum_l\lVert q_l^I-q_l^T\rVert_2^2.
\end{aligned}}
\label{eq:mixture-quadratic}
\end{equation}
Here $a^\star$ is the unconstrained minimizer for $\kappa>0$, and its projection
onto $[0,1]$ is the feasible minimizer; $\kappa=0$ gives a constant curve
(Appendix~\ref{sup:ratio-law}).

Equation~\eqref{eq:mixture-quadratic} separates the minimum imbalance from
curvature $\kappa$ around that minimum. A large $\kappa$ creates a narrow
low-imbalance region, while reducing the modality gap flattens the curve. For
varying token ratios with $R(a^\star)$ and $a^\star$ held fixed, reducing
$\kappa$ lowers the cost of both ratio variation and a mean ratio away from
$a^\star$ (Appendix~\ref{sup:ratio-law}).

\paragraph{Connecting the law to physical inputs.}
Resolution and tiling change both token counts and conditional load profiles. We
evaluate $\sqrt{R(a)}$ at each workload's measured token ratio and compare it
with a new forward pass on the same requests (Sec.~\ref{sec:exp-mixture}), where
the residual captures changes in the modality profiles
(Appendix~\ref{sup:profile-drift}).

\subsection{Routing Structure and Aggregation}
\label{sec:phenomenon-geometry}
Figure~\ref{fig:method}(b) shows image and text tokens in different
router-input regions, and the measured expert preferences in panel (e) follow
this modality structure. With the load-response law, these findings motivate
separate modality losses.

Separating modality losses also requires choosing how to group tokens within
each modality. In router-input space, Fig.~\ref{fig:method}(f) plots the sample
cosine gap, the mean similarity within a sample minus that across samples,
averaging $0.019$ for images and $0.036$ for text on the split backbone (native
diagnostics in Appendix~\ref{sup:geometry-detail}). Within-image tokens are thus
more similar than cross-image tokens, and varying patch counts let large images
dominate flat pooling. \method{} therefore weights image profiles equally and
pools valid text tokens within each microbatch (Sec.~\ref{sec:exp-ablation}).

\section{\method{}}
\label{sec:method}

\subsection{Separate Modality Objectives}
\label{sec:method-separation}

To prevent cross-modal cancellation, \method{} uses separate image and text
balance terms (Fig.~\ref{fig:method}(e,f)) with the Switch hard--soft product
form. When hard and soft modality means coincide, each term penalizes the
squared CV of its mean profile (identity and gradients in
Appendix~\ref{sup:conditional-identifiability}). Using only router outputs and
group IDs, this objective is independent of expert initialization.

\subsection{Image Profiles and Pooled Text}
\label{sec:method-formulation}

Within each MoE layer (layer index $l$ suppressed), let a nonempty token group
$m$ contain $S_m$ valid tokens. For images, each group $m\in\mathcal M_I$ is
one source image; since $S_m$ varies with resolution and tiling, flat token
pooling would let patch-rich images dominate. Using $f_{t,i}$ and $p_{t,i}$
from Eq.~\eqref{eq:laux}, \method{} averages hard dispatch and soft gate
profiles separately within each group:
\begin{equation}
P_{m,i}=\frac1{S_m}\sum_{t\in m}p_{t,i},\qquad
F_{m,i}=\frac1{S_m}\sum_{t\in m}f_{t,i}.
\label{eq:intra}
\end{equation}
For a nonempty group collection $\mathcal M$ in the local microbatch, it
averages these profiles with equal weight per group:
\begin{equation}
\bar P_i=\frac1{|\mathcal M|}\sum_{m\in\mathcal M}P_{m,i},\qquad
\bar F_i=\frac1{|\mathcal M|}\sum_{m\in\mathcal M}F_{m,i},
\label{eq:inter}
\end{equation}
and computes one balance loss from the means (with $\Linst(\varnothing)=0$):
\begin{equation}
\Linst(\mathcal M)=N\sum_{i=1}^{N}\bar F_i\bar P_i.
\label{eq:linst}
\end{equation}
Write $\bar F^r,\bar P^r$ for the means over $\mathcal M_r$ ($r\in\{I,T\}$).
Unlike token-weighted image load $q_l^I$, $\bar F^I$ weights every image in
$\mathcal M_I$ equally while letting individual images favor different experts
(Fig.~\ref{fig:method}(g)). For text, all valid text tokens in the microbatch
form one pooled group $T_{\mathrm{pool}}$, so
$\mathcal M_T=\{T_{\mathrm{pool}}\}$ when text is present and
$\mathcal M_T=\varnothing$ otherwise. Pooling text tokens gives equal weight per
token rather than overweighting short requests. The per-layer objective is
\begin{equation}
\begin{aligned}
\Lreba={}&\lambda_I\Linst(\mathcal M_I)\\
&+\lambda_T\Linst(\mathcal M_T).
\end{aligned}
\label{eq:lreba}
\end{equation}
Equal-image weights act within $\Linst(\mathcal M_I)$, while $\lambda_I$ and
$\lambda_T$ weight the two modality losses.

\subsection{Implementation}
Let $T_I,T_T$ count valid image and text tokens in the local microbatch,
excluding padding. For $T_I+T_T>0$, we set $\lambda_I=T_I/(T_I+T_T)$ and
$\lambda_T=T_T/(T_I+T_T)$, with absent modalities contributing zero. Weighting by realized token shares keeps per-token gradient scale consistent
across varying batch compositions: fixed equal weights would over-penalize
sparse text tokens in vision-heavy batches, whereas instance-count weights
skew balance as image counts vary (Sec.~5.4). Each
device computes statistics over its local microbatch, averages $\Lreba$ across
all $L$ MoE layers, and adds $\lambda_{\mathrm{aux}}$ times that mean to the
task loss before gradient accumulation. Hard counts are stop-gradient
statistics, while soft profiles supply router gradients. \method{} adds only
group aggregation and preserves router and expert parameter counts
(Appendix~\ref{sup:reba-algorithm}).

\section{Experiments}
\label{sec:experiments}

\subsection{Setup}
\label{sec:exp-setup}

\paragraph{Models and training protocol.}
Our primary model is \textbf{Split-Qwen3VL-4B}, built from Qwen3-VL-4B-Instruct
by splitting each of its 36 SwiGLU FFN layers into $N{=}4$ disjoint experts with
top-$2$ routing \cite{zhang2022moefication,zhu2024llamamoe} at an unchanged FFN
parameter budget. We apply this construction to \textbf{Split-Qwen2.5VL-3B},
\textbf{Split-Qwen2VL-7B}, and \textbf{Split-InternVL3-8B}, plus an eight-expert
top-$4$ Split-Qwen3VL-4B. Within each backbone and expert count, all runs start
from the same split checkpoint and share one-epoch supervised fine-tuning on
Cambrian-737K \cite{tong2024cambrian} with fixed architecture, top-$k$ routing,
and execution. Across all four backbones, \textsc{No-Aux} anchors unregularized
routing and \stdaux{} \cite{fedus2022switch} compares pooled balancing against
\method{}, using a seven-point coefficient sweep for \stdaux{} and \method{} on
a fixed 500-request probe (466 image-bearing requests;
Appendices~\ref{sup:implementation} and~\ref{sup:coefficient-protocol}).
On Split-Qwen3VL-4B, we also compare against a \textbf{LIMoE-style}
modality-entropy loss \cite{mustafa2022limoe} to test whether anti-collapse
regularization stabilizes load under composition shifts
(Appendix~\ref{sup:additional}).

\paragraph{Metrics.}
All load metrics use true top-$k$ dispatch counts. Tables~\ref{tab:main}
and~\ref{tab:ablation} report mean layer CV, $L^{-1}\sum_l\mathrm{CV}(q_l)$.
Workload studies and the load columns of Table~\ref{tab:objective-quality}
report aggregate $\mathrm{RMSCV}=\sqrt{L^{-1}\sum_l\mathrm{CV}^2(q_l)}$, whose
square is $R(a)$ in Sec.~\ref{sec:phenomenon-law}. Resolution and tiling sweeps
also summarize curves by normalized AUC over the observed image-token fraction
$a$. Task quality uses VLMEvalKit scorers \cite{duan2024vlmevalkit}.

\begin{figure*}[t]
\centering
\includegraphics[width=0.48\textwidth]{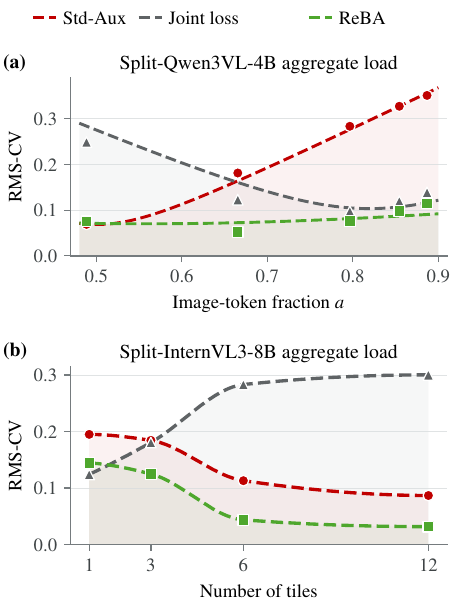}\hfill
\includegraphics[width=0.48\textwidth]{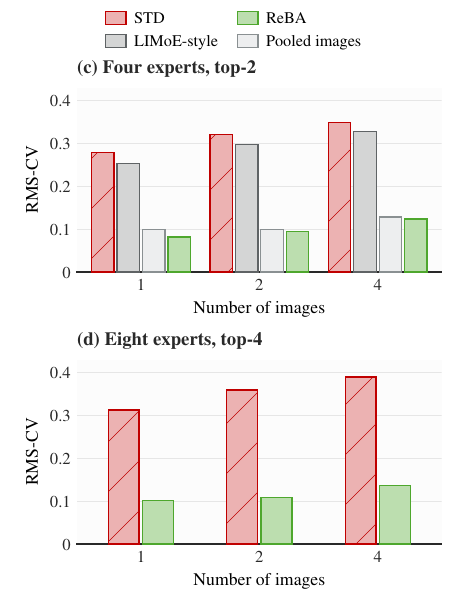}
\caption{From predicted load response to physical visual workloads.
(a) Qwen resolution sweep: dashed curves are fixed-profile predictions and
markers are measured runs. (b) InternVL tiling: markers are measured loads and
dashed lines connect tested settings.
(c,d) Measured loads for one, two, and four distinct images on Split-Qwen3VL-4B
with four experts/top-$2$ (c) and eight experts/top-$4$ (d). STD is red, ReBA is green, and gray marks the LIMoE-style baseline in (c)
and the internal ablations (\textsc{Joint loss} and
\textsc{Pooled images}) analyzed in Sec.~\ref{sec:exp-ablation}.
Appendices~\ref{sup:additional} and~\ref{sup:workload shifts} give full
protocols and numerical tables.}
\label{fig:mixture-robustness}
\end{figure*}

\subsection{Predicting and Stabilizing Load Across Workloads}
\label{sec:exp-mixture}

We first test whether the reference image and text profiles from one condition
predict how a fixed router responds to real resolution changes. In
Fig.~\ref{fig:mixture-robustness}(a), dashed curves evaluate $\sqrt{R(a)}$ from
one reference condition on Split-Qwen3VL-4B without refitting, while markers
show physical forward passes across five resolutions
(Table~\ref{sup:tab:qwen-correspondence}). Measured loads closely track the
predictions, with mean absolute RMS-CV errors of $0.006$ (\stdaux{}) and
$0.013$ (\method{}).

Shrinking the modality gap during training flattens this load response across
workloads. Across the five Qwen resolutions
(Fig.~\ref{fig:mixture-robustness}(a)), \stdaux{} RMS-CV rises to $0.351$,
whereas \method{} stays below $0.115$ and lowers 95th-percentile request RMS-CV
at every setting (Table~\ref{sup:tab:qwen-tails}). On Split-InternVL3-8B,
varying dynamic tiling from one to twelve tiles
(Fig.~\ref{fig:mixture-robustness}(b)) keeps \method{} below \stdaux{} at every
tile count, cutting twelve-tile RMS-CV from $0.087$ to $0.032$. Over the
observed image-token fractions, \method{} lowers normalized AUC by $65.2\%$ on
Qwen and by $37.0\%$ on InternVL (Table~\ref{tab:mixture-robustness}).
Diagnostics on two released native MoEs in Appendix~\ref{sup:native} also
reveal large image--text load gaps and high-resolution profile drift, showing
that this modality gap is not limited to split models.

\begin{table*}[!t]
\caption{Task scores and benchmark-input load. Each cell reports \emph{task score ($\uparrow$) / mean layer CV ($\downarrow$)}. Mean layer CV uses true top-$k$ dispatch counts. Bold marks the best value among all three methods within each backbone (ties included). MME reports its raw total and is excluded from the six-benchmark mean task score (all other scores in \%), while mean CV averages all seven benchmarks.}
\label{tab:main}
\centering
\small
\setlength{\tabcolsep}{0pt}
\renewcommand{\arraystretch}{1.0}
\begin{tabular*}{\linewidth}{@{}p{.13\linewidth}*{8}{>{\centering\arraybackslash}p{.10875\linewidth}}@{}}
\toprule
Method & POPE & \shortstack{Hallusion\\Bench} & MME & MMBench & MMStar & SEEDBench & ScienceQA & Avg. \\
\midrule
\multicolumn{9}{@{}l}{\textbf{Split-Qwen3VL-4B}} \\
\textsc{No-Aux} & 86.9\,/\,0.73 & \textbf{66.9}\,/\,0.67 & 1597\,/\,0.70 & 63.5\,/\,0.67 & 43.6\,/\,0.69 & 70.9\,/\,0.70 & 64.3\,/\,0.66 & 66.0\,/\,0.69 \\
\stdaux{} & 83.2\,/\,0.47 & 65.2\,/\,0.45 & 1520\,/\,0.48 & \textbf{65.7}\,/\,0.37 & 45.3\,/\,0.42 & 71.2\,/\,0.42 & \textbf{64.4}\,/\,0.37 & 65.8\,/\,0.43 \\
\rowcolor{rebatint}
\textbf{\method{}} & \textbf{87.8}\,/\,\textbf{0.12} & 63.1\,/\,\textbf{0.25} & \textbf{1768}\,/\,\textbf{0.18} & 65.5\,/\,\textbf{0.13} & \textbf{46.1}\,/\,\textbf{0.13} & \textbf{71.5}\,/\,\textbf{0.12} & 62.3\,/\,\textbf{0.20} & \textbf{66.1}\,/\,\textbf{0.16} \\
\midrule
\multicolumn{9}{@{}l}{\textbf{Split-Qwen2.5VL-3B}} \\
\textsc{No-Aux} & \textbf{87.5}\,/\,0.76 & 61.5\,/\,0.68 & 1637\,/\,0.69 & 52.9\,/\,0.68 & 42.7\,/\,0.71 & 64.8\,/\,0.72 & 61.8\,/\,0.67 & 61.9\,/\,0.70 \\
\stdaux{} & 86.9\,/\,0.35 & 59.7\,/\,0.26 & \textbf{1717}\,/\,0.27 & \textbf{58.0}\,/\,0.20 & \textbf{44.7}\,/\,0.24 & \textbf{67.3}\,/\,0.25 & 61.7\,/\,0.22 & 63.1\,/\,0.26 \\
\rowcolor{rebatint}
\textbf{\method{}} & 85.2\,/\,\textbf{0.15} & \textbf{64.0}\,/\,\textbf{0.25} & 1696\,/\,\textbf{0.21} & 57.4\,/\,\textbf{0.19} & 43.9\,/\,\textbf{0.17} & 67.1\,/\,\textbf{0.16} & \textbf{62.0}\,/\,\textbf{0.21} & \textbf{63.3}\,/\,\textbf{0.19} \\
\midrule
\multicolumn{9}{@{}l}{\textbf{Split-Qwen2VL-7B}} \\
\textsc{No-Aux} & \textbf{87.4}\,/\,0.70 & \textbf{64.8}\,/\,0.63 & 1879\,/\,0.65 & \textbf{69.8}\,/\,0.64 & \textbf{47.8}\,/\,0.65 & \textbf{72.0}\,/\,0.66 & \textbf{69.4}\,/\,0.64 & \textbf{68.5}\,/\,0.65 \\
\stdaux{} & 86.4\,/\,0.23 & 60.6\,/\,0.24 & 1925\,/\,0.23 & 67.9\,/\,0.18 & 46.7\,/\,0.18 & 71.6\,/\,0.17 & 67.8\,/\,0.21 & 66.8\,/\,0.21 \\
\rowcolor{rebatint}
\textbf{\method{}} & 87.3\,/\,\textbf{0.10} & 60.1\,/\,\textbf{0.21} & \textbf{1936}\,/\,\textbf{0.19} & 67.2\,/\,\textbf{0.15} & 47.1\,/\,\textbf{0.14} & 71.8\,/\,\textbf{0.13} & 66.0\,/\,\textbf{0.19} & 66.6\,/\,\textbf{0.16} \\
\midrule
\multicolumn{9}{@{}l}{\textbf{Split-InternVL3-8B}} \\
\textsc{No-Aux} & 87.2\,/\,0.70 & \textbf{56.0}\,/\,0.64 & 1980\,/\,0.67 & \textbf{70.5}\,/\,0.63 & \textbf{51.1}\,/\,0.66 & 73.9\,/\,0.66 & \textbf{73.3}\,/\,0.63 & \textbf{68.7}\,/\,0.66 \\
\stdaux{} & \textbf{88.0}\,/\,0.25 & 55.2\,/\,0.21 & 1906\,/\,0.25 & 70.4\,/\,0.14 & 50.8\,/\,0.18 & 72.8\,/\,0.19 & 72.2\,/\,0.16 & 68.2\,/\,0.20 \\
\rowcolor{rebatint}
\textbf{\method{}} & 86.6\,/\,\textbf{0.16} & 55.4\,/\,\textbf{0.18} & \textbf{2027}\,/\,\textbf{0.18} & 69.6\,/\,\textbf{0.11} & \textbf{51.1}\,/\,\textbf{0.14} & \textbf{74.2}\,/\,\textbf{0.14} & 72.5\,/\,\textbf{0.13} & 68.2\,/\,\textbf{0.15} \\
\bottomrule
\end{tabular*}
\end{table*}

\begin{table*}[!t]
\caption{Additional balancing methods (a) and eight-expert scaling (b) on Split-Qwen3VL-4B. Under a common runtime, task columns ($\uparrow$) report POPE F1, HallusionBench aAcc, raw MME sum, 0--100 OCRBench, and percentages elsewhere (subscores in Table~\ref{sup:tab:additional-quality}(b)). Load columns give aggregate RMS-CV ($\downarrow$) on the separate 500-request probe at 1, 2, and 4 images per request with fixed checkpoint, text, \mbox{and per-image preprocessing.} Bold marks best per block (ties included).}
\label{tab:objective-quality}
\centering\small
\setlength{\tabcolsep}{0pt}
\renewcommand{\arraystretch}{1.0}
\begin{tabular*}{\linewidth}{@{}p{.18\linewidth}*{2}{>{\raggedleft\arraybackslash}p{.057\linewidth}}>{\raggedleft\arraybackslash}p{.072\linewidth}*{8}{>{\raggedleft\arraybackslash}p{.057\linewidth}}*{3}{>{\raggedleft\arraybackslash}p{.059333\linewidth}}@{}}
\toprule
& \multicolumn{11}{c}{Task scores $\uparrow$} & \multicolumn{3}{c}{RMS-CV $\downarrow$} \\
\cmidrule(lr){2-12}\cmidrule(l){13-15}
Method & POPE & \shortstack{Hallu.\\Bench} & MME & \shortstack{MM\\Bench} & \shortstack{MM\\Star} & \shortstack{SEED\\Bench} & \shortstack{Sci.\\QA} & \shortstack{Text\\VQA} & \shortstack{Doc\\VQA} & \shortstack{Chart\\QA} & \shortstack{OCR\\Bench} & \shortstack{1\\image} & \shortstack{2\\images} & \shortstack{4\\images} \\
\midrule
\multicolumn{15}{@{}l}{\textbf{(a) Additional balancing methods}} \\
\stdaux{} & 83.45 & \textbf{65.62} & 1530.20 & \textbf{65.89} & 44.93 & 71.14 & \textbf{64.65} & 69.39 & 78.44 & \textbf{68.24} & 62.70 & 0.279 & 0.320 & 0.349 \\
LIMoE-style & \textbf{87.69} & 65.40 & 1691.59 & 65.29 & \textbf{47.27} & 71.30 & 63.61 & \textbf{69.51} & \textbf{80.48} & 67.92 & 64.20 & 0.253 & 0.298 & 0.328 \\
\rowcolor{rebatint}
\textbf{\method{}} & 87.62 & 63.51 & \textbf{1831.89} & \textbf{65.89} & 46.20 & \textbf{71.64} & 63.11 & 69.31 & 78.69 & 67.84 & \textbf{64.30} & \textbf{0.082} & \textbf{0.094} & \textbf{0.124} \\
\midrule
\multicolumn{15}{@{}l}{\textbf{(b) More experts (8 experts, top-$4$)}} \\
\stdaux{} & \textbf{88.52} & \textbf{62.46} & \textbf{1820.65} & \textbf{67.10} & 44.73 & 70.77 & \textbf{66.73} & 68.73 & \textbf{80.16} & 67.56 & 63.00 & 0.312 & 0.359 & 0.390 \\
\rowcolor{rebatint}
\textbf{\method{}} & 86.43 & 60.36 & 1795.86 & 64.52 & \textbf{45.13} & \textbf{70.90} & 64.45 & \textbf{69.02} & 79.33 & \textbf{68.84} & \textbf{64.60} & \textbf{0.101} & \textbf{0.109} & \textbf{0.136} \\
\bottomrule
\end{tabular*}
\end{table*}

\begin{figure*}[t]
\centering
\includegraphics[width=\textwidth]{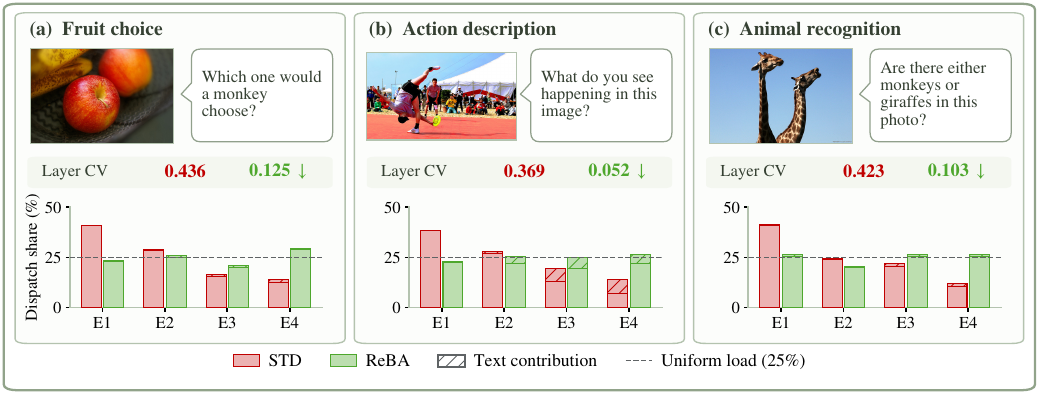}
\caption{Expert load on three real image--question inputs. STD is red and
ReBA is green; hatching denotes text dispatch. Arrows mark lower CV than STD.
All panels show layer 19 of 36
with identical axes and uniform load marked by a dashed line. Counts use the
complete recorded conversations; Appendix~\ref{sup:qualitative} gives the
sample indices and measurement protocol.}
\label{fig:qualitative-case}
\end{figure*}

\ifdefined\ReBAArxiv
\begin{table*}[!t]
\caption{Objective ablation on four-expert Split-Qwen3VL-4B, reporting mean layer CV ($\downarrow$) from true top-$k$ dispatches on the 500-request probe. A check (\cmark) uses the column's design choice, whereas a cross (\xmark) uses its alternative: one coupled loss (rows~a--c), token-pooled images (rows~a,f), equal-weight text requests (rows~b,e), or row~c's matched weights (row~d). A dash (`--`) marks coupled rows without separate modality weights (Appendix~\ref{sup:ablation-design}). Bold marks the lowest CV per column.}
\label{tab:ablation}
\centering\small
\setlength{\tabcolsep}{0pt}
\begin{tabular*}{\linewidth}{@{}p{.25\linewidth}*{2}{>{\centering\arraybackslash}p{.12\linewidth}}>{\centering\arraybackslash}p{.09\linewidth}>{\centering\arraybackslash}p{.13\linewidth}*{3}{>{\centering\arraybackslash}p{.096666\linewidth}}@{}}
\toprule
& \multicolumn{4}{c}{Design choices} & \multicolumn{3}{c}{Mean layer CV $\downarrow$} \\
\cmidrule(lr){2-5}\cmidrule(l){6-8}
Variant & \shortstack{Separate\\modality losses} & \shortstack{Equal-image\\averaging} & \shortstack{Pooled\\text} & \shortstack{Token-share\\loss weights} & Image & Text & Overall \\
\midrule
(a) \stdaux{} & \xmark & \xmark & \cmark & -- & 0.447 & 0.445 & 0.273 \\
(b) \textsc{Joint + per-request text} & \xmark & \cmark & \xmark & -- & 0.567 & 0.543 & 0.348 \\
(c) \textsc{Joint loss} & \xmark & \cmark & \cmark & -- & 0.170 & 0.640 & 0.097 \\
\midrule
(d) \textsc{Matched weights} & \cmark & \cmark & \cmark & \xmark & 0.109 & 0.477 & 0.154 \\
(e) \textsc{Per-request text} & \cmark & \cmark & \xmark & \cmark & 0.108 & 0.138 & 0.098 \\
(f) \textsc{Pooled images} & \cmark & \xmark & \cmark & \cmark & 0.111 & \textbf{0.108} & 0.092 \\
\midrule
\rowcolor{rebatint}
(g) \textbf{\method{}} & \cmark & \cmark & \cmark & \cmark & \textbf{0.099} & 0.110 & \textbf{0.077} \\
\bottomrule
\end{tabular*}
\end{table*}

\begin{table*}[!t]
\begin{minipage}[t]{.48\textwidth}
{
\centering
\captionof{table}{Overall mean layer CV ($\downarrow$) across coefficients $\lambda_{\mathrm{aux}}$ on four-expert top-$2$ Split-Qwen3VL-4B, from true top-$k$ dispatches on the fixed 500-request probe. Bold marks the lowest CV.}
\label{tab:coefficient-main}
{\small
\setlength{\tabcolsep}{0pt}
\begin{tabular*}{\linewidth}{@{}p{.19\linewidth}*{7}{>{\centering\arraybackslash}p{.115714\linewidth}}@{}}
\toprule
$\lambda_{\mathrm{aux}}$ & 0.001 & 0.002 & 0.005 & 0.010 & 0.020 & 0.050 & 0.100 \\
\midrule
\stdaux{} & .523 & .468 & .421 & .375 & .339 & .321 & .273 \\
\rowcolor{rebatint}
\method{} & \textbf{.492} & \textbf{.463} & \textbf{.383} & \textbf{.245} & \textbf{.158} & \textbf{.105} & \textbf{.077} \\
\bottomrule
\end{tabular*}
}
}

\end{minipage}\hfill
\begin{minipage}[t]{.48\textwidth}
{
\centering
\captionof{table}{Within-image load and between-profile variation on the fixed probe.
$\rho_2$ is the share of mean per-profile squared imbalance explained by
between-profile variation.}
\label{tab:mechanism}

{\small
\setlength{\tabcolsep}{0pt}
\begin{tabular*}{\linewidth}{@{}p{.18\linewidth}>{\centering\arraybackslash}p{.30\linewidth}*{2}{>{\centering\arraybackslash}p{.26\linewidth}}@{}}
\toprule
Method & \shortstack{Within-image\\CV $\downarrow$} &
\shortstack{Image share\\$\rho_2$} &
\shortstack{Text share\\$\rho_2$} \\
\midrule
\stdaux{} & 0.48 & 0.19 & 0.30 \\
\rowcolor{rebatint}
\method{} & \textbf{0.21} & \textbf{0.79} & \textbf{0.87} \\
\bottomrule
\end{tabular*}
}

}

\end{minipage}
\end{table*}

\fi

\subsection{Joint Load Balance and Task Quality}
\label{sec:exp-pareto}
\label{sec:exp-generality}

\paragraph{Cross-backbone results.}
Table~\ref{tab:main} evaluates load balance and task accuracy across four
vision--language backbones and seven benchmarks. \method{} lowers mean layer CV
relative to \stdaux{} on every benchmark and backbone, cutting average CV from
$0.43$ to $0.16$ on Split-Qwen3VL-4B and to $0.15$--$0.19$ on the other three
models while keeping mean task scores within $0.3$ points of \stdaux{}.

\paragraph{Additional balancing methods.}
Table~\ref{tab:objective-quality}(a) and Fig.~\ref{fig:mixture-robustness}(c)
compare \stdaux{}, LIMoE-style balancing \cite{mustafa2022limoe}, and \method{}.
Beyond 11 standard benchmarks, the 500-request probe tests fixed checkpoints at
1, 2, and 4 images per request. Adding images raises the visual token share to
test balance under composition shift. At four images, probe RMS-CV is $0.349$
for \stdaux{} and $0.328$ for LIMoE-style, while \method{} cuts it to $0.124$
with competitive \mbox{quality across all 11 benchmarks.}

\paragraph{More experts.}
Table~\ref{tab:objective-quality}(b) and Fig.~\ref{fig:mixture-robustness}(d)
evaluate eight-expert top-$4$ routing. As request images increase from one to
four on the 500-request probe, \stdaux{} RMS-CV rises from $0.312$ to $0.390$,
whereas \method{} stays between $0.101$ and $0.136$. Across the 11 benchmarks,
\method{} also maintains competitive overall quality at eight experts.

\subsection{Decomposing the \method{} Objective}
\label{sec:exp-ablation}
\label{sec:exp-additional}

\ifdefined\ReBAArxiv\else

\fi

Table~\ref{tab:ablation} and Fig.~\ref{fig:mixture-robustness}(a,c) test three
questions on four-expert Split-Qwen3VL-4B: whether instance grouping can
replace modality separation, what separating the modality losses isolates under
matched weights, and which aggregation unit works best within each modality
(Appendix~\ref{sup:ablation-design}).

\paragraph{Modality separation and token-share weighting.}
Grouping instances inside a single coupled loss (rows~b,c) does not prevent
cross-modal cancellation. In \textsc{Joint loss} (row~c), overall CV drops
to $0.097$ at the probe ratio, but opposing errors cancel across modalities,
leaving text CV at $0.640$ and a steep U-shaped resolution curve in
Fig.~\ref{fig:mixture-robustness}(a) ($\kappa=0.639$). Comparing row~c with
\textsc{Matched weights} (row~d) holds the within-modality weights fixed and
removes only the cross-modal products from the loss. This cuts image CV to
$0.109$ and text CV to $0.477$, showing that row~c's low mixed CV hid severe
conditional skew. Switching from those image-count-dependent weights to realized
token shares (row~d to row~g) then cuts text CV from $0.477$ to $0.110$ and
overall CV to $0.077$ ($\kappa=0.031$).

\paragraph{Within-modality aggregation.}
With separate modality losses and token-share weights fixed, comparing
\textsc{Per-request text} (row~e) with \method{} (row~g) isolates text grouping.
Averaging profiles per text request gives tokens in short prompts more weight,
raising text CV from $0.110$ to $0.138$ and overall CV from $0.077$ to $0.098$.
Comparing \textsc{Pooled images} (row~f) with \method{} (row~g) isolates
image grouping. Averaging image profiles equally in the visual loss lowers
image CV from $0.111$ to $0.099$ and overall CV from $0.092$ to $0.077$, and
cuts RMS-CV at all three image counts in
Fig.~\ref{fig:mixture-robustness}(c) (full results in
Table~\ref{sup:tab:additional-quality}(a)).

\paragraph{Effect of balancing strength.}
We vary overall balancing strength $\lambda_{\mathrm{aux}}$ across seven weights
from $0.001$ to $0.1$ on the 500-request probe
(Table~\ref{tab:coefficient-main}). \method{} achieves lower overall mean layer
CV than \stdaux{} at all seven coefficients, including $0.077$ versus $0.273$ at
$\lambda_{\mathrm{aux}}=0.1$ (conditional CVs in
Table~\ref{sup:tab:coefficient}).

\subsection{What \method{} Routing Looks Like}
\label{sec:exp-mechanism}
\label{sec:exp-routing}
\label{sec:exp-systems}

Table~\ref{tab:mechanism} shows how balancing the mean image profile reshapes
individual profiles on the fixed probe (Appendix~\ref{sup:image-routes}).
\method{} lowers mean within-image CV from $0.48$ to $0.21$ while raising the
between-profile share $\rho_2$ from $0.19$ to $0.79$ for images and from $0.30$
to $0.87$ for text. Individual images thus use experts more evenly on average,
while a larger share of the remaining profile variation offsets across the
batch, matching Fig.~\ref{fig:method}(g).

\paragraph{Individual request routing (Fig.~\ref{fig:qualitative-case}).}
At layer~19, three inputs show how opposing preferences
(Sec.~\ref{sec:phenomenon-shortcut}) skew single requests. Under \stdaux{},
images pile onto $\mathrm{E1}$ and $\mathrm{E2}$ while text favors
$\mathrm{E3}$ and $\mathrm{E4}$, skewing load toward the visual stream.
\method{} routes all three evenly, cutting panel~(b)'s busiest-expert share
from $38.6\%$ to $26.6\%$ (uniform: $25\%$). Since expert-parallel layers sync
on the busiest expert, idealized bottleneck work $T^{\mathrm{proxy}} =
\textstyle\sum_l \max_i n_{l,i}$ under uniform placement
(Appendix~\ref{sup:request-systems}) is $1.23$--$1.25\times$ higher for
\stdaux{} than \method{} at the three largest Qwen resolutions
(Table~\ref{sup:tab:idealized-compute}), cutting straggler delay.

\section{Conclusion}
\label{sec:conclusion}

We show that pooled MoE balancing hides \emph{cross-modal load cancellation},
where opposing image and text errors offset at one mix until composition shifts
break balance. Under fixed modality profiles, we derive an exact quadratic law
showing that one reference run predicts this response, with sensitivity governed
by the modality gap. Guided by this law, we propose \method{} to shrink this gap
during fine-tuning, lowering imbalance across four split backbones and diverse
visual workloads at comparable accuracy. While our training study uses split
backbones and analytical bottleneck work, large modality gaps in released native
MoEs motivate scaling \method{} to native pre-training and live expert-parallel
serving.

\clearpage
{\small
\bibliographystyle{ieeenat_fullname}
\bibliography{refs}
}

\clearpage
\appendix
\setcounter{table}{0}
\setcounter{figure}{0}
\setcounter{algorithm}{0}
\setcounter{equation}{0}
\renewcommand{\thetable}{S\arabic{table}}
\renewcommand{\thefigure}{S\arabic{figure}}
\renewcommand{\thealgorithm}{S\arabic{algorithm}}
\renewcommand{\theequation}{S\arabic{equation}}
\renewcommand{\theHtable}{S\arabic{table}}
\renewcommand{\theHfigure}{S\arabic{figure}}
\renewcommand{\theHequation}{S\arabic{equation}}
\providecommand{\theHalgorithm}{}
\renewcommand{\theHalgorithm}{S\arabic{algorithm}}
\section{Implementation and Evaluation Protocol}
\label{sup:implementation}

\subsection{ReBA Computation}
\label{sup:reba-algorithm}

Algorithm~\ref{sup:alg:reba} summarizes the \method{} computation for one MoE
layer and one device-local microbatch \mbox{using the notation of Sec.~\ref{sec:method}.}

\begin{algorithm}[t]
\caption{\method{} (Relax Within, Balance Across)}
\label{sup:alg:reba}
\small
\begin{algorithmic}[1]
\REQUIRE router outputs $\{p_{t,i}\},\{f_{t,i}\}$, group sets $\mathcal M_I,\mathcal M_T$, token counts $T_I,T_T$
\STATE $\Lreba\gets 0$
\IF{$T_I+T_T=0$}
  \RETURN $\Lreba$
\ENDIF
\STATE $\lambda_I\gets T_I/(T_I+T_T)$;\quad $\lambda_T\gets T_T/(T_I+T_T)$
\FOR{$r\in\{I,T\}$}
  \IF{$\mathcal M_r\neq\varnothing$}
    \STATE compute $P_m,F_m$ for each $m\in\mathcal M_r$ via Eq.~\eqref{eq:intra}
    \STATE $\bar P^r\gets|\mathcal M_r|^{-1}\sum_{m\in\mathcal M_r}P_m$
    \STATE $\bar F^r\gets|\mathcal M_r|^{-1}\sum_{m\in\mathcal M_r}F_m$
    \STATE $\Lreba\gets\Lreba+\lambda_r N\sum_{i=1}^{N}\bar F_i^r\bar P_i^r$
  \ENDIF
\ENDFOR
\RETURN $\Lreba$
\end{algorithmic}
\end{algorithm}

A token-to-image membership mask contracts token routing statistics into
per-image means $P_m,F_m$ and averages them across the $M=|\mathcal M_I|$
images in the microbatch, leaving the router and expert parameter tensors
unchanged. On the same tokens, flat pooling weights image $m$ by
$S_m/\sum_{m'\in\mathcal M_I}S_{m'}$, whereas \method{} assigns each image
weight $1/M$; when all images have equal token count $S_m$, these means
coincide ($\bar F^I=q_l^I$). The visual loss is computed once from $\bar F^I$
and $\bar P^I$ rather than by averaging per-image losses. For text, pooling
gives every valid token weight $1/T_T$ ($\bar F^T=q_l^T$ on the same tokens),
whereas equal weighting across $B_T$ nonempty text requests assigns a token in
request $j$ of length $S_j$ weight $1/(B_T S_j)$.

\subsection{Dense-to-MoE Construction and Training}
\label{sup:split-construction}
\label{sup:training}
To build each split MoE backbone, we partition the intermediate neurons of
every dense SwiGLU FFN into $N$ disjoint groups. Expert $i$ receives the
matching rows of $W_{\rm gate}$ and $W_{\rm up}$ and columns of $W_{\rm down}$,
which preserves the dense FFN parameter budget. Our primary model,
Split-Qwen3VL-4B, uses four experts with top-$2$ routing at all 36 decoder
layers. We apply this construction to Qwen2.5-VL-3B, Qwen2-VL-7B, and InternVL3-8B, and also test an eight-expert top-$4$ Split-Qwen3VL-4B.

Within each backbone, \textsc{No-Aux}, \stdaux{}, and \method{} start from the
same split checkpoint and share all task-training settings over one epoch on
Cambrian-737K \cite{tong2024cambrian}. The vision encoder is frozen while the
language model, routers, and unfrozen multimodal modules are trained. Primary
4B training uses bfloat16, FlashAttention, learning rate $10^{-5}$, a cosine
schedule, warmup ratio $0.03$, maximum sequence length $4096$, device batch
size four, and 16 gradient-accumulation steps, giving a recorded global batch
of 512 on NVIDIA H20 GPUs with ZeRO-1 and gradient checkpointing. We evaluate the final checkpoint after one full epoch, where the extended objective and eight-expert training runs each complete 1,440 steps.

\subsection{Probe, Metrics, and Workloads}
\label{sup:routing-protocol}
\label{sup:provenance}
The fixed routing probe is a seeded reservoir sample of 500 rows from the
cleaned Cambrian training corpus (seed 42), including 466 image-bearing rows.
Under Qwen preprocessing with both pixel bounds fixed at $1{,}003{,}520$, the
default probe contains $469{,}224$ image tokens and $118{,}218$ text tokens,
giving image-token fraction $a=0.7988$. Qwen forwards process the full recorded conversation, where each valid token contributes $k$ hard dispatches per MoE layer and profiles are normalized by total dispatches.

As defined in the main paper, Tables~\ref{tab:main} and~\ref{tab:ablation}
report mean layer CV, whereas the load columns of
Table~\ref{tab:objective-quality} and the workload studies use aggregate RMS-CV.
Workload composition is the image-token fraction of the complete evaluated
request set. For $J$ evaluated conditions sorted by image-token fraction
$a_1<\cdots<a_J$ with aggregate RMS-CVs $y_1,\ldots,y_J$, the normalized area is
\[
\operatorname{AUC}=
\frac{\sum_{j=1}^{J-1}(a_{j+1}-a_j)(y_j+y_{j+1})/2}{a_J-a_1},
\]
and minimum, worst, and range summarize the same stated interval. Request p90
and p95 first compute RMS-CV per request and then take percentiles. Confidence
intervals use 1,000 paired bootstrap resamples of request IDs, reaggregating
expert counts for each resample.

Resolution and tiling sweeps evaluate the same 466 image-bearing requests at
every setting. Qwen pixel budgets are $200{,}704$, $401{,}408$, $802{,}816$,
$1{,}204{,}224$, and $1{,}605{,}632$, while InternVL uses one, three, six, or
twelve tiles with its standard chat template. Image-count tests retain all 500 probe rows, assigning one, two, or four distinct images cyclically from the probe's ordered unique image paths to each image-bearing request while holding the text prompt and per-image pixel limits fixed.

Task quality is evaluated on standard benchmark inputs with VLMEvalKit
\cite{duan2024vlmevalkit}, using rule-based or exact-match scoring and no GPT
judge, with preprocessing and evaluation matched within each comparison.
Table~\ref{tab:main} reports the four-backbone suite, and
Table~\ref{tab:objective-quality} reports the 11-benchmark Split-Qwen3VL-4B
results under a common runtime (corresponding component scores in
Table~\ref{sup:tab:additional-quality}(b)).

\subsection{Concrete Examples}
\label{sup:qualitative}
Figure~\ref{fig:qualitative-case} uses three single-turn, image-bearing records
from the fixed probe (zero-based indices 112, 209, and 285), chosen for clear
subjects and readable questions before inspecting their routing measurements.
All panels show layer 19 of 36, where each bar records an expert's share of all
top-$2$ dispatches across the complete recorded question--answer conversation.
Solid and hatched segments separate image and text tokens, and each panel
displays the input image, question stem, and layer-19 CV.

\section{Proofs and Load Identities}
\label{sup:proofs}

\subsection{Composition-Shift Law}
\label{sup:ratio-law}
We first derive how squared imbalance $R(a)$ varies with the image-token
fraction $a$ under fixed conditional profiles. Let $d_l=q_l^I-q_l^T$ and
$u=(1/N,\ldots,1/N)$. Expanding $q_l(a)=q_l^T+a d_l$ gives
\begin{align}
R(a)&=\frac NL\sum_l\|q_l^T-u+a d_l\|_2^2\\
&=\kappa a^2+2ba+c,\label{sup:eq:quadratic}
\end{align}
where $\kappa=(N/L)\sum_l\|d_l\|_2^2$,
$b=(N/L)\sum_l d_l^\top(q_l^T-u)$, and
$c=(N/L)\sum_l\|q_l^T-u\|_2^2$.
For $\kappa>0$, completing the square yields $a^\star=-b/\kappa$ and
$R(a)=R(a^\star)+\kappa(a-a^\star)^2$. The feasible minimum projects $a^\star$
onto $[0,1]$; $\kappa=0$ gives a constant curve. If $q_l(a_0)=u$ at every
layer, then $q_l(a)-u=(a-a_0)d_l$ and $\sqrt{R(a)}=|a-a_0|\sqrt\kappa$.

For workload image-token fraction $\mathsf A$ with $\mu=\mathbb E[\mathsf A]$
and $\sigma^2=\operatorname{Var}(\mathsf A)$, taking expectations of the
completed-square form and using
$\mathbb E[(\mathsf A-a^\star)^2]
=\operatorname{Var}(\mathsf A)+(\mathbb E[\mathsf A]-a^\star)^2$ yields
\begin{equation}
\mathbb E[R(\mathsf A)]
=R(a^\star)+\kappa\bigl[\sigma^2+(\mu-a^\star)^2\bigr].
\label{eq:expected-composition}
\end{equation}
Ratio variation adds $\kappa\sigma^2$ to expected squared imbalance, and a mean
ratio away from $a^\star$ adds $\kappa(\mu-a^\star)^2$. With $R(a^\star)$ and
$a^\star$ held fixed, reducing $\kappa$ lowers both terms. Applying
$\|x-y\|_2^2\le2\|x-u\|_2^2+2\|y-u\|_2^2$ further gives
\[
\kappa\le\frac2L\sum_l[\mathrm{CV}^2(q_l^I)+\mathrm{CV}^2(q_l^T)],
\]
so conditional balance bounds the curvature $\kappa$.

\paragraph{Physical profile changes.}
\label{sup:profile-drift}
With reference conditional profiles $q_l^I,q_l^T$, let preprocessing setting $s$
have image-token fraction $a_s$, conditional profiles $q_{s,l}^I,q_{s,l}^T$,
predicted load $q_{\rm pred,l}=a_s q_l^I+(1-a_s)q_l^T$, and physical load
$q_{\rm phys,l}=a_s q_{s,l}^I+(1-a_s)q_{s,l}^T$. Writing
$r_l=a_s(q_{s,l}^I-q_l^I)+(1-a_s)(q_{s,l}^T-q_l^T)$ gives
$q_{\rm phys,l}=q_{\rm pred,l}+r_l$, so the squared aggregate RMS-CVs
$R_{\rm pred}=R(a_s)$ and $R_{\rm phys}=L^{-1}\sum_l\mathrm{CV}^2(q_{\rm phys,l})$
satisfy
\begin{equation}
\begin{aligned}
R_{\rm phys}={}&R_{\rm pred}+\frac NL\sum_l\|r_l\|_2^2\\
&+\frac{2N}{L}\sum_l\langle q_{\rm pred,l}-u,r_l\rangle.
\end{aligned}
\label{sup:eq:physical-square}
\end{equation}
The last two terms separate the magnitude of profile change from its
interaction with the predicted load error, explaining the physical residuals in
Appendix~\ref{sup:workload shifts}.

\subsection{Separate Modality Objectives}
\label{sup:conditional-identifiability}

We next show why a mixed loss admits opposing modality errors and how separate
terms penalize conditional imbalance. For sparse routing ($k<N$) and $0<a<1$,
choose a nonzero zero-sum vector $\Delta$ small enough that $q^I=u+\Delta$ and
$q^T=u-a\Delta/(1-a)$ are feasible top-$k$ dispatch profiles. Then
$a q^I+(1-a)q^T=u$, so a mixed loss admits conditionally imbalanced solutions
at uniform load.

For the modality mean profiles $\bar F^r,\bar P^r$ ($r\in\{I,T\}$), let
$\delta\bar F^r=\bar F^r-u$ and $\delta\bar P^r=\bar P^r-u$. The surrogate
expands as
\begin{equation}
N(\bar F^r)^\top\bar P^r=1+N(\delta\bar F^r)^\top\delta\bar P^r.
\label{sup:eq:hard-soft}
\end{equation}
Under aligned profiles ($\bar P^r=\bar F^r$), ReBA's excess becomes
\[
\Lreba-(\lambda_I+\lambda_T)
=\lambda_I\mathrm{CV}^2(\bar F^I)+\lambda_T\mathrm{CV}^2(\bar F^T).
\]
With positive weights, its minimum occurs at $\bar F^I=\bar F^T=u$, and an
excess at most $\epsilon\ge0$ implies $\mathrm{CV}^2(\bar F^r)\le\epsilon/\lambda_r$.
Table~\ref{sup:tab:alignment} reports empirical directional agreement between
token-pooled hard and soft modality profiles on the fixed probe. Treating hard counts as stop-gradient, the gradient of
$\Linst(\mathcal M_r)$ with respect to router logit $g_i(x_t)$ for token
$t\in m$ ($m\in\mathcal M_r$) is
\[
\begin{aligned}
\frac{\partial\Linst(\mathcal M_r)}{\partial g_i(x_t)}
&=\frac{N}{|\mathcal M_r|S_m}\,p_{t,i}
\left(\bar F_i^r-\sum_{j=1}^N p_{t,j}\bar F_j^r\right),
\end{aligned}
\]
so an expert above the token's soft load average receives a downward update
within each modality.

\subsection{Within-Image and Between-Image Balance}
\label{sup:image-routes}
Finally, we decompose per-image imbalance into batch-mean imbalance and
between-image variation. For $M=|\mathcal M_I|$ normalized image profiles $F_m$
($m\in\mathcal M_I$) with mean $\bar F=\bar F^I=M^{-1}\sum_mF_m$, expanding
around $\bar F$ and using $\sum_m(F_m-\bar F)=0$ gives
\begin{equation}
\frac1M\sum_m\|F_m-u\|_2^2
=\|\bar F-u\|_2^2+\frac1M\sum_m\|F_m-\bar F\|_2^2,
\label{sup:eq:within-between}
\end{equation}
where the second term also equals $(2M^2)^{-1}\sum_{m,n}\|F_m-F_n\|_2^2$.
Multiplying by $N$ gives the squared-CV decomposition, and the between-profile
share is
\[
\rho_2=
\frac{M^{-1}\sum_m\mathrm{CV}^2(F_m)-\mathrm{CV}^2(\bar F)}
{M^{-1}\sum_m\mathrm{CV}^2(F_m)}.
\]
For a positive denominator, $\rho_2\in[0,1]$ (undefined when all profiles are
uniform). Table~\ref{tab:mechanism} averages layerwise $\rho_2$ and mean
within-group CV across layers to separate individual profile concentration from
mean balance, using nonempty text requests as the text groups.

\section{Objective Comparisons and Expert-Work Analysis}
\label{sup:additional-results}

\subsection{Coefficient Selection and Routing Alignment}
\label{sup:coefficient-protocol}
\label{sup:training-alignment}
To select the auxiliary weight fairly, we train \stdaux{} and \method{}
separately at each coefficient from $0.001$ to $0.1$ and evaluate the final
checkpoint on the fixed 500-request probe. While
Table~\ref{tab:coefficient-main} in the main paper reports overall mean layer
CV, Table~\ref{sup:tab:coefficient} provides the image and text conditional
CVs. Within \method{}, the inner modality weights remain the realized token
shares. The outer coefficient $0.1$ yields the lowest conditional and overall
load in this grid for both methods and is used in our main comparisons.

\begin{table}[!htbp]
\caption{Image and text conditional mean layer CV ($\downarrow$) across auxiliary coefficients $\lambda_{\mathrm{aux}}$ on the fixed 500-request probe, complementing the overall loads in Table~\ref{tab:coefficient-main}. Each entry uses true top-$k$ dispatch counts from a separate training run.}
\label{sup:tab:coefficient}
\centering\small
\setlength{\tabcolsep}{0pt}
\begin{tabular*}{\linewidth}{@{}p{.28\linewidth}*{4}{>{\centering\arraybackslash}p{.18\linewidth}}@{}}
\toprule
& \multicolumn{2}{c}{\stdaux{}} & \multicolumn{2}{c}{\method{}} \\
\cmidrule(lr){2-3}\cmidrule(l){4-5}
$\lambda_{\mathrm{aux}}$ & Image & Text & Image & Text \\
\midrule
0.001 & .604 & .646 & .572 & .623 \\
0.002 & .586 & .628 & .534 & .588 \\
0.005 & .550 & .567 & .458 & .564 \\
0.010 & .547 & .593 & .303 & .442 \\
0.020 & .537 & .535 & .200 & .362 \\
0.050 & .519 & .506 & .130 & .204 \\
0.100 & .447 & .445 & .099 & .110 \\
\bottomrule
\end{tabular*}
\end{table}

Table~\ref{sup:tab:alignment} reports the layerwise cosine similarity between
centered token-pooled hard dispatch and soft gate profiles for each modality on
the fixed probe. Across both modalities, these centered routing profiles show
predominantly positive directional agreement across MoE layers.

\begin{table}[!htbp]
\caption{Hard--soft routing alignment on the fixed probe. Median and interquartile range summarize layerwise cosines between token-pooled hard and soft deviations from uniform routing for each modality; the last column counts layers with positive cosine.}
\label{sup:tab:alignment}

\centering
\small
\setlength{\tabcolsep}{4pt}
\begin{tabular}{llcc}
\toprule
Method & Modality & Median [IQR] & Positive layers \\
\midrule
\stdaux{} & Image & $0.969\ [0.938,0.982]$ & $36/36$ \\
\stdaux{} & Text  & $0.953\ [0.916,0.979]$ & $36/36$ \\
\method{} & Image & $0.770\ [0.380,0.886]$ & $31/36$ \\
\method{} & Text  & $0.831\ [0.582,0.901]$ & $29/36$ \\
\bottomrule
\end{tabular}

\end{table}

\subsection{Matched Objectives and Expert Counts}
\label{sup:additional}
On Split-Qwen3VL-4B, we conduct three complementary evaluations: an additional
balancing-method comparison (\stdaux{}, LIMoE-style, and \method{} with four
experts and top-$2$ routing), an eight-expert top-$4$ scaling comparison
(\stdaux{} and \method{}), and the internal \textsc{Pooled images} ablation.
To compare against modality-specific entropy regularization
\cite{mustafa2022limoe} under the same generative VLM training task, our
LIMoE-style baseline combines mixed Switch balancing with local entropy and
per-modality global entropy hinges. Local weights are $0$ for images and $0.1$
for text, both hinge weights are $0.1$, global thresholds are $\log4$ for
images and $\log2$ for text, and the outer coefficient is $0.1$. Each
modality's mean router distribution is computed within the training microbatch,
and the global hinge averages its entropy across MoE layers. Eight-expert
\stdaux{} and \method{} use top-$4$ routing to preserve the active expert
fraction, while \textsc{Pooled images} retains \method{}'s separate modality
terms, token-share weights, and pooled text but pools visual tokens across
images. All six runs evaluate the final checkpoint on the same 500-request
probe.

\paragraph{Ablation objective definitions.}
\label{sup:ablation-design}
With $\mathcal M_I$, $\mathcal M_T=\{T_{\mathrm{pool}}\}$, and means
$\bar F^r,\bar P^r$ ($r\in\{I,T\}$) defined in Sec.~\ref{sec:method}, let
$\mathcal M_{\mathrm{req}}$ denote the set of $B_T=|\mathcal M_{\mathrm{req}}|$
nonempty text requests in the microbatch. In Table~\ref{tab:ablation},
\textsc{Joint + per-request text} (row~b) forms $M+B_T$ equally weighted groups
($M=|\mathcal M_I|$) in one coupled loss
$\Linst(\mathcal M_I\cup\mathcal M_{\mathrm{req}})$. \textsc{Joint loss}
(row~c) forms $M+1$ groups in $\Linst(\mathcal M_I\cup\mathcal M_T)$, giving
combined profiles $\bar F^{\mathrm{all}}=\frac{M}{M+1}\bar F^I+\frac{1}{M+1}\bar F^T$
and $\bar P^{\mathrm{all}}=\frac{M}{M+1}\bar P^I+\frac{1}{M+1}\bar P^T$.
Expanding $N(\bar F^{\mathrm{all}})^\top\bar P^{\mathrm{all}}$ yields
within-modality terms
$\frac{M^2}{(M+1)^2}\Linst(\mathcal M_I)+\frac{1}{(M+1)^2}\Linst(\mathcal M_T)$
plus cross-modal products. \textsc{Matched weights} (row~d) drops only the
cross-modal products while keeping $\lambda_I=M^2/(M+1)^2$ and
$\lambda_T=1/(M+1)^2$. Rows~(e--g) use token-share weights
$\lambda_I=T_I/(T_I+T_T)$ and $\lambda_T=T_T/(T_I+T_T)$:
\textsc{Per-request text} (row~e) replaces $\mathcal M_T$ with
$\mathcal M_{\mathrm{req}}$, and \textsc{Pooled images} (row~f) pools all valid
\mbox{image tokens into one visual group.}

\begin{table}[!tp]
\caption{Multi-image probe settings on Split-Qwen3VL-4B. Each condition evaluates all 500 requests with fixed text and per-image preprocessing, reporting the realized image-token fraction $a$ from processed tokens (RMS-CV results in Tables~\ref{tab:objective-quality} and~\ref{sup:tab:additional-quality}(a)).}
\label{sup:tab:all-workloads}
\centering\small
\setlength{\tabcolsep}{0pt}
\begin{tabular*}{\linewidth}{@{\extracolsep{\fill}}lrr@{}}
\toprule
Images per request & Requests & Image fraction \\
\midrule
1 & 500 & 0.799 \\
2 & 500 & 0.887 \\
4 & 500 & 0.939 \\
\bottomrule
\end{tabular*}
\end{table}

\begin{table*}[!tp]
\caption{Extended evaluations on Split-Qwen3VL-4B under the common runtime. (a) Full 11-benchmark task scores ($\uparrow$, same metrics and scales as Table~\ref{tab:objective-quality}) and 500-request probe aggregate RMS-CV ($\downarrow$) for the \textsc{Pooled images} internal ablation, for comparison with \method{} in Table~\ref{tab:objective-quality}(a). (b) POPE and HallusionBench percentages and raw MME component subscores ($\uparrow$) for the balancing methods, the \textsc{Pooled images} ablation, and eight-expert scaling. \mbox{Bold marks the best score within the method and eight-expert comparison groups (ties included).}}
\label{sup:tab:additional-quality}
\centering\small
\setlength{\tabcolsep}{0pt}
\textbf{(a) Image-aggregation ablation}\par\smallskip
\begin{tabular*}{\linewidth}{@{}p{.18\linewidth}*{2}{>{\raggedleft\arraybackslash}p{.057\linewidth}}>{\raggedleft\arraybackslash}p{.072\linewidth}*{8}{>{\raggedleft\arraybackslash}p{.057\linewidth}}*{3}{>{\raggedleft\arraybackslash}p{.059333\linewidth}}@{}}
\toprule
& \multicolumn{11}{c}{Task scores $\uparrow$} & \multicolumn{3}{c}{RMS-CV $\downarrow$} \\
\cmidrule(lr){2-12}\cmidrule(l){13-15}
Method & POPE & \shortstack{Hallu.\\Bench} & MME & \shortstack{MM\\Bench} & \shortstack{MM\\Star} & \shortstack{SEED\\Bench} & \shortstack{Sci.\\QA} & \shortstack{Text\\VQA} & \shortstack{Doc\\VQA} & \shortstack{Chart\\QA} & \shortstack{OCR\\Bench} & \shortstack{1\\image} & \shortstack{2\\images} & \shortstack{4\\images} \\
\midrule
Pooled images & 87.06 & 64.88 & 1726.78 & 65.64 & 45.73 & 71.46 & 64.15 & 69.79 & 80.24 & 67.24 & 64.60 & 0.098 & 0.099 & 0.128 \\
\bottomrule
\end{tabular*}
\par\medskip
\textbf{(b) Component scores}\par\smallskip
\begin{tabular*}{\linewidth}{@{\extracolsep{\fill}}lrrrrrr@{}}
\toprule
& \multicolumn{3}{c}{Additional methods} & \multicolumn{1}{c}{Ablation} & \multicolumn{2}{c}{8 experts} \\
\cmidrule(lr){2-4}\cmidrule(lr){5-5}\cmidrule(lr){6-7}
Metric & STD & LIMoE-style & ReBA & Pooled images & STD & ReBA \\
\midrule
POPE accuracy & 85.48 & 88.26 & \textbf{88.36} & 88.09 & \textbf{88.96} & 87.61 \\
HallusionBench fAcc & 31.21 & 32.08 & \textbf{32.37} & 32.37 & 32.08 & \textbf{32.95} \\
HallusionBench qAcc & 38.02 & \textbf{38.24} & 36.92 & 37.36 & \textbf{36.26} & 33.19 \\
MME perception & 1231.98 & 1349.09 & \textbf{1428.68} & 1439.28 & 1396.37 & \textbf{1417.29} \\
MME reasoning & 298.21 & 342.50 & \textbf{403.21} & 287.50 & \textbf{424.29} & 378.57 \\
\bottomrule
\end{tabular*}
\end{table*}

Table~\ref{sup:tab:all-workloads} lists the multi-image probe settings and
realized image-token fractions used in Fig.~\ref{fig:mixture-robustness}(c,d).
In the main paper, Table~\ref{tab:objective-quality} pairs those 1-, 2-, and
4-image RMS-CVs with primary task scores across all 11 benchmarks under a
common runtime for the additional balancing methods (panel~a) and eight-expert
scaling (panel~b). Table~\ref{sup:tab:additional-quality}(a) reports the
matching 11-benchmark scores and multi-image RMS-CVs for the \textsc{Pooled
images} internal ablation, and Table~\ref{sup:tab:additional-quality}(b)
provides the POPE, HallusionBench, and MME component subscores across all
evaluated runs.

\subsection{Busiest-Expert Work}
\label{sup:request-systems}
To relate dispatch balance to idealized expert bottleneck work, let $n_{l,i}$
denote the dispatch count of expert $i$ at layer $l$, and define
\[
T^{\rm proxy}=\sum_l\max_i n_{l,i},\qquad
U=\frac{\sum_lN^{-1}\sum_i n_{l,i}}{T^{\rm proxy}}.
\]
This proxy models expert-only work under equally fast experts and ideal
placement, leaving communication and non-MoE layers outside the model. Because
matched requests have equal total dispatches, the proxy work ratio satisfies
$T^{\rm proxy}_{\rm STD}/T^{\rm proxy}_{\rm ReBA}=U_{\rm ReBA}/U_{\rm STD}$.
Table~\ref{sup:tab:idealized-compute} reports this ratio and paired
request-bootstrap intervals across every physical workload setting. At the three largest Qwen settings, ReBA cuts bottleneck work by $1.23$--$1.25\times$ and improves utilization at all four InternVL tile counts.

\begin{table}[!htbp]
\caption{Busiest-expert work across physical shifts on 466 paired requests. $U$ is utilization in the expert-work model of Appendix~\ref{sup:request-systems}; $S_{\mathrm{ideal}}=T_{\stdauxsub}^{\mathrm{proxy}}/T_{\mathrm{ReBA}}^{\mathrm{proxy}}$. Ratios above one favor ReBA. Intervals use 1,000 paired request-bootstrap resamples.}
\label{sup:tab:idealized-compute}

\centering
\small
\setlength{\tabcolsep}{2.5pt}
\begin{tabular}{lrrrr}
\toprule
Setting & $U_{\stdauxsub}$ & $U_{\mathrm{ReBA}}$ &
$S_{\mathrm{ideal}}$ & Paired 95\% interval \\
\midrule
\multicolumn{5}{l}{\textit{Split-Qwen3VL-4B}} \\
$200{,}704$ px   & 0.930 & 0.914 & 0.983 & [0.973, 0.998] \\
$401{,}408$ px   & 0.815 & 0.938 & 1.151 & [1.133, 1.167] \\
$802{,}816$ px   & 0.740 & 0.912 & 1.233 & [1.222, 1.245] \\
$1{,}204{,}224$ px & 0.712 & 0.888 & 1.247 & [1.236, 1.257] \\
$1{,}605{,}632$ px & 0.698 & 0.871 & 1.248 & [1.238, 1.257] \\
\midrule
\multicolumn{5}{l}{\textit{Split-InternVL3-8B}} \\
1 tile  & 0.809 & 0.846 & 1.046 & [1.043, 1.049] \\
3 tiles & 0.810 & 0.864 & 1.066 & [1.061, 1.071] \\
6 tiles & 0.883 & 0.948 & 1.074 & [1.069, 1.078] \\
12 tiles & 0.914 & 0.966 & 1.057 & [1.053, 1.060] \\
\bottomrule
\end{tabular}

\end{table}

\section{Complete Visual Workload Measurements}
\label{sup:workload shifts}

\subsection{Curve Summaries}
\label{sup:fixed-landscapes}
Table~\ref{tab:mixture-robustness} summarizes the controlled composition curves
over $a\in[0.1,0.9]$ and the physical resolution and tiling sweeps over their
observed image-token fractions. Normalized AUC averages RMS-CV across each
stated interval, and $\kappa$ is the squared-imbalance curvature from
Eq.~\eqref{eq:mixture-quadratic}. ReBA achieves the lowest worst load and AUC
in both studies on each backbone.
\begin{table}[!htbp]
\caption{Load response across compositions. Controlled curves use $a\in[0.1,0.9]$; physical sweeps use their observed image-token fractions. AUC averages RMS-CV over the stated interval, and $\kappa$ is the squared-imbalance curvature. Lower is better.}
\label{tab:mixture-robustness}

\centering
\small
\textbf{(a) Controlled recomposition}\par\smallskip
\setlength{\tabcolsep}{2.5pt}
\begin{tabular}{@{}lrrrr@{}}
\toprule
Method & Min & Worst & AUC & $\kappa$ \\
\midrule
\multicolumn{5}{l}{\textit{Split-Qwen3VL-4B}} \\
\stdaux{} & \textbf{.0687} & .3695 & .1998 & .8205 \\
Joint loss  & .1034 & .5842 & .2939 & .6390 \\
Matched weights & .1305 & .4510 & .2790 & .2188 \\
Per-request text  & .1011 & .1350 & .1102 & \textbf{.0265} \\
\method{} & .0702 & \textbf{.1067} & \textbf{.0811} & .0306 \\
\midrule
\multicolumn{5}{l}{\textit{Split-InternVL3-8B}} \\
\stdaux{} & .1917 & .4152 & .2585 & .5160 \\
\method{} & \textbf{.1444} & \textbf{.2331} & \textbf{.1692} & \textbf{.1059} \\
\bottomrule
\end{tabular}
\par\medskip
\textbf{(b) Real resolution / tiling sweeps}\par\smallskip
\setlength{\tabcolsep}{3pt}
\begin{tabular}{@{}lrrr@{}}
\toprule
Method & Worst & AUC & Range \\
\midrule
\multicolumn{4}{l}{\textit{Split-Qwen3VL-4B}} \\
\stdaux{} & .3511 & .2041 & .2821 \\
Joint loss & .2491 & .1453 & .1516 \\
\method{}& \textbf{.1148} & \textbf{.0711} & \textbf{.0623} \\
\midrule
\multicolumn{4}{l}{\textit{Split-InternVL3-8B}} \\
\stdaux{} & .1949 & .1626 & \textbf{.1082} \\
Joint loss  & .3004 & .2022 & .1763 \\
\method{} & \textbf{.1444} & \textbf{.1024} & .1128 \\
\bottomrule
\end{tabular}

\end{table}

\subsection{Resolution and Tiling}
\label{sup:qwen-physical}
\label{sup:internvl-tiling}
Table~\ref{sup:tab:qwen-correspondence} lists the five physical Qwen resolution
runs in Fig.~\ref{fig:mixture-robustness}(a) and evaluates the fixed-profile
prediction from the $1{,}003{,}520$-pixel reference at each setting's measured
fraction $a$. Across all 15 method--setting pairs, the mean absolute residual between physical and predicted RMS-CV is $0.014$ ($0.006$ for STD and $0.013$ for ReBA).

\begin{table}[!htbp]
\caption{Qwen resolution sweep on 466 paired requests. Predictions evaluate the fixed-profile curve at each measured image-token fraction $a$, using the $1{,}003{,}520$-pixel reference. Residual is physical minus predicted RMS-CV.}
\label{sup:tab:qwen-correspondence}

\centering
\small
\setlength{\tabcolsep}{3pt}
\begin{tabular}{lrrrr}
\toprule
Method & $a$ & Prediction & Physical & Residual \\
\midrule
\stdaux{} & 0.4885 & 0.0696 & 0.0690 & $-0.0006$ \\
 & 0.6654 & 0.1641 & 0.1813 & $+0.0172$ \\
 & 0.7965 & 0.2765 & 0.2836 & $+0.0071$ \\
 & 0.8543 & 0.3275 & 0.3274 & $-0.0001$ \\
 & 0.8868 & 0.3563 & 0.3511 & $-0.0052$ \\
\midrule
\textsc{Joint loss} & 0.4885 & 0.2840 & 0.2491 & $-0.0349$ \\
 & 0.6654 & 0.1607 & 0.1232 & $-0.0376$ \\
 & 0.7965 & 0.1050 & 0.0975 & $-0.0076$ \\
 & 0.8543 & 0.1071 & 0.1197 & $+0.0126$ \\
 & 0.8868 & 0.1166 & 0.1390 & $+0.0224$ \\
\midrule
\method{} & 0.4885 & 0.0713 & 0.0755 & $+0.0042$ \\
 & 0.6654 & 0.0726 & 0.0526 & $-0.0201$ \\
 & 0.7965 & 0.0815 & 0.0765 & $-0.0050$ \\
 & 0.8543 & 0.0871 & 0.0985 & $+0.0115$ \\
 & 0.8868 & 0.0906 & 0.1148 & $+0.0243$ \\
\bottomrule
\end{tabular}

\end{table}

Table~\ref{sup:tab:internvl-correspondence} gives the InternVL tiling
measurements from Fig.~\ref{fig:mixture-robustness}(b) alongside predictions
from the one-tile conditional profiles. Increasing the tile count changes the
conditional profiles in addition to shifting $a$: at twelve tiles, physical
RMS-CV is $0.087$ for STD and $0.032$ for ReBA, compared with fixed-profile
predictions of $0.307$ and $0.172$. Equation~\eqref{sup:eq:physical-square} separates this drift from the token-ratio effect.

\begin{table}[!htbp]
\caption{InternVL tiling sweep on 466 paired requests. Predictions use the one-tile conditional profiles at each measured fraction $a$. Residual is physical minus predicted RMS-CV.}
\label{sup:tab:internvl-correspondence}

\centering
\small
\setlength{\tabcolsep}{2pt}
\begin{tabular}{@{}llrrrr@{}}
\toprule
Method & Tiles & $a$ & Pred. & Phys. & Resid. \\
\midrule
\stdaux{} & 1  & .6617 & .1949 & .1949 & $+.0000$ \\
             & 3  & .7840 & .2279 & .1844 & $-.0435$ \\
             & 6  & .9243 & .2948 & .1130 & $-.1818$ \\
             & 12 & .9461 & .3068 & .0868 & $-.2201$ \\
\midrule
\method{}    & 1  & .6617 & .1444 & .1444 & $+.0000$ \\
             & 3  & .7840 & .1498 & .1249 & $-.0248$ \\
             & 6  & .9243 & .1677 & .0437 & $-.1241$ \\
             & 12 & .9461 & .1715 & .0316 & $-.1398$ \\
\bottomrule
\end{tabular}

\end{table}

\paragraph{Request-level load.}
Table~\ref{sup:tab:qwen-tails} reports 90th and 95th request RMS-CV percentiles
on the same 466 paired requests. ReBA lowers both p90 and p95 at every tested resolution, extending the aggregate balance improvement to the upper tail of individual request loads.

\begin{table}[!htbp]
\caption{Upper-tail request load across Qwen resolutions. Percentiles use per-request RMS-CV on the 466 paired requests.}
\label{sup:tab:qwen-tails}

\centering
\small
\setlength{\tabcolsep}{3pt}
\begin{tabular}{lrrrr}
\toprule
& \multicolumn{2}{c}{\stdaux{}} & \multicolumn{2}{c}{\method{}} \\
\cmidrule(lr){2-3}\cmidrule(lr){4-5}
Pixel budget & p90 & p95 & p90 & p95 \\
\midrule
$200{,}704$ & .406 & .438 & .351 & .386 \\
$401{,}408$ & .443 & .462 & .322 & .367 \\
$802{,}816$ & .469 & .492 & .310 & .352 \\
$1{,}204{,}224$ & .480 & .499 & .310 & .349 \\
$1{,}605{,}632$ & .485 & .504 & .319 & .358 \\
\bottomrule
\end{tabular}

\end{table}

\subsection{Training Composition and the Load Minimum}
\label{sup:training-composition}
To compare the \stdaux{} load minimum on Split-Qwen3VL-4B with its fine-tuning token
mix, we reconstruct training token counts on CPU from Cambrian-737K. We uniformly
sample $20{,}000$ records without replacement and apply the training tokenizer, chat
template, and adaptive image preprocessing. Across this sample, the aggregate
image-token fraction is $0.538$. Forming random device-local microbatches of four and
averaging the batch fraction over $100$ reshuffles of the sample gives a mean of
$0.551$.

On the same Split-Qwen3VL-4B \stdaux{} checkpoint shown in Fig.~\ref{fig:teaser}, the
reference-profile curve reaches its minimum at $a^\star=0.501$. This minimum lies
$3.7$--$5.0$ percentage points below the two reconstructed training fractions. As an
observational check on Split-Qwen3VL-4B, this comparison places the \stdaux{} load
minimum near typical reconstructed training compositions.

\section{Routing Structure and Native MoE Diagnostics}
\label{sup:native}
\label{sup:geometry-detail}

Inference studies on pretrained multimodal MoEs report modality-dependent
expert hotspots under varying visual-to-text workloads \cite{li2026macs}. To
check whether the modality load gaps from the main paper also occur in two
released native MoEs, we inspect router-input structure, image--text load gaps,
and profile drift across image resolutions.

\subsection{Modality and Image Structure}
\begin{figure*}[t]
\centering
\includegraphics[width=0.98\textwidth]{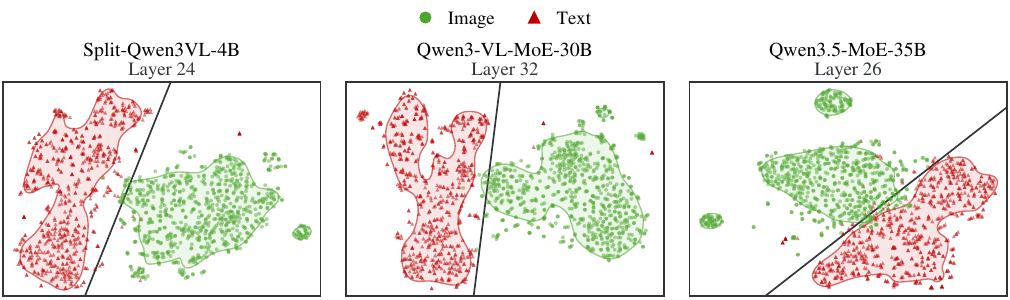}
\includegraphics[width=0.98\textwidth]{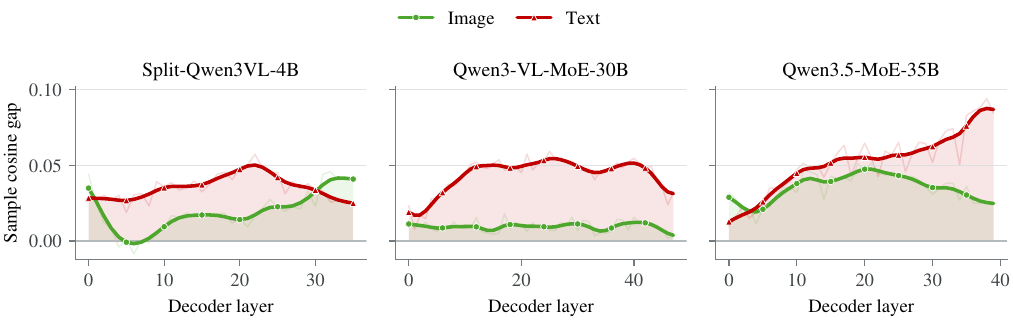}
\caption{Router-input structure on the split backbone and two native MoEs.
Top: image (green) and text (red) t-SNE embeddings, with separators as visual
guides. Bottom: same-minus-different sample cosine gaps in representation
space. Faint traces give layer values; solid curves are Gaussian-smoothed
trends. The split-model column provides a common reference for the native
models.}
\label{fig:routing-geometry}
\end{figure*}

To check whether the router-input structure seen on our split backbone also
appears in pretrained multimodal MoEs, we inspect router-input states from the
no-aux split backbone and two released native MoEs in
Fig.~\ref{fig:routing-geometry} (where the split-model column reproduces
Fig.~\ref{fig:method}(b,f)). The top row projects image and text router inputs
into t-SNE coordinates, while the bottom row plots the sample cosine gap in the
original representation space. For images, this gap is the mean cosine
similarity within an image minus the mean across different images; for text,
sample identity is the source request. These diagnostics motivate the modality
and image aggregation units evaluated in the main paper.

\subsection{Native Conditional Loads}
\label{sup:native-protocol}
\label{sup:native-source}
We next measure conditional dispatch loads in the two released native models.
Qwen3-VL-MoE-30B-A3B-Instruct has 48 MoE layers, 128 routed experts, and
top-$8$ dispatch, while Qwen3.5-MoE-35B-A3B has 40 layers, 256 experts, and
top-$8$ dispatch, with shared experts excluded. Qwen3.5 supplies selected
expert indices directly, and Qwen3-VL's top-$8$ logits recover dispatch.
Vision-boundary, video, padding, and special control tokens are excluded when computing the text routing statistics.

The native load study evaluates 64 paired Cambrian requests across three
resolution settings: Low uses pixel bounds $(65{,}536,200{,}704)$, Source uses
the default lower bound and maximum $1{,}003{,}520$, and High fixes both bounds
at $1{,}605{,}632$. These settings produce $11{,}844$, $17{,}751$, and
$103{,}267$ image tokens, while text counts stay fixed at $10{,}978$ for 30B
and $11{,}284$ for 35B.

\begin{table}[!htbp]
\caption{Native model loads on the 64-request source probe: conditional and mixed RMS-CV, centered-profile orientation, conditional gap, curvature, and effective expert counts.}
\label{sup:tab:native-source}

\centering
\small
\setlength{\tabcolsep}{2pt}
\begin{tabular}{@{}lcccccc@{}}
\toprule
& & \multicolumn{3}{c}{RMS-CV} & & \\
\cmidrule(lr){3-5}
Model & $N/k$ & Image & Text & Mixed & Cos. & Neg. \\
\midrule
30B & $128/8$ & .822 & 1.051 & .632 & $-.075$ & 66.7\% \\
35B & $256/8$ & .997 & 1.242 & .820 & $+.108$ & 22.5\% \\
\bottomrule
\end{tabular}
\par\smallskip
\begin{tabular}{@{}lccc@{}}
\toprule
Model & $G_2$ & $\kappa$ & Effective $I/T$ \\
\midrule
30B & .121 & 1.865 & 76.3/61.3 \\
35B & .094 & 2.246 & 130.9/98.3 \\
\bottomrule
\end{tabular}

\end{table}

Table~\ref{sup:tab:native-source} shows substantial conditional load gaps in
both native models. Here Cos.\ is the median layerwise cosine between
$q_l^I-u$ and $q_l^T-u$, Neg.\ is the fraction of layers with negative cosine,
and Effective $I/T$ is the median across layers of the per-layer effective
expert count $1/\sum_{i=1}^N(q_{l,i}^r)^2$ ($r\in\{I,T\}$). Finally,
$G_2=\sqrt{L^{-1}\sum_l\lVert q_l^I-q_l^T\rVert_2^2}$ is the RMS conditional
$\ell_2$ gap and $\kappa=L^{-1}\sum_l\kappa_l=NG_2^2$. Layerwise, the
curvature $\kappa_l=N\lVert q_l^I-q_l^T\rVert_2^2$ expands as
\[
\begin{aligned}
\kappa_l={}&\mathrm{CV}^2(q_l^I)+\mathrm{CV}^2(q_l^T)\\
&-2\mathrm{CV}(q_l^I)\mathrm{CV}(q_l^T)\cos\theta_l,
\end{aligned}
\]
where $\theta_l$ is the angle between $q_l^I-u$ and $q_l^T-u$. The magnitude
and orientation terms average $1.780$ and $+0.084$ for 30B, and $2.536$ and
$-0.290$ for 35B, so conditional magnitudes dominate $\kappa$ in both models.

\paragraph{Count-matched reference.}
\label{sup:native-null}
In 10,000 uniform-routing trials, every token selects $k$ distinct experts
uniformly while preserving token counts, expert count, and layer count. Every
observed conditional or mixed RMS-CV in Table~\ref{sup:tab:native-null} exceeds
its count-matched null mean by over $23\times$.

\begin{table}[!htbp]
\caption{Observed native loads versus a count-matched uniform-routing reference. The null mean uses 10,000 trials preserving token counts, $N$, top-$k$, and layer count.}
\label{sup:tab:native-null}

\centering
\small
\setlength{\tabcolsep}{4pt}
\begin{tabular}{llrrr}
\toprule
Model & Profile & Observed & Null mean & Ratio \\
\midrule
30B & Image & .822 & .0291 & 28.3 \\
    & Text  & 1.051 & .0370 & 28.4 \\
    & Mixed & .632 & .0229 & 27.7 \\
\midrule
35B & Image & .997 & .0418 & 23.8 \\
    & Text  & 1.242 & .0524 & 23.7 \\
    & Mixed & .820 & .0327 & 25.1 \\
\bottomrule
\end{tabular}

\end{table}

\subsection{Native Resolution Changes}
\label{sup:native-physical}
\label{sup:native-drift}
Table~\ref{sup:tab:native-physical} compares physical RMS-CV across the three
native resolution settings with predictions from the Source conditional
profiles and 95\% paired request-bootstrap intervals. At High resolution, both
physical loads fall below the fixed-profile prediction, and the 35B model
reverses its predicted increase. These measured residuals illustrate the profile-drift terms in Eq.~\eqref{sup:eq:physical-square} on native MoEs.

\FloatBarrier
\begin{center}
\begin{minipage}{\columnwidth}
\captionof{table}{Native model resolution sweeps on 64 paired requests. Predictions use the source conditional profiles; residual is physical minus predicted RMS-CV. Intervals are 95\% paired request-bootstrap intervals for physical load.}
\label{sup:tab:native-physical}

\centering
\small
\setlength{\tabcolsep}{2pt}
\textit{Qwen3-VL-MoE-30B}\par\smallskip
\begin{tabular}{@{}lrrrrl@{}}
\toprule
Setting & $a$ & Pred. & Phys. & Resid. & Phys. 95\% CI \\
\midrule
Low    & .519 & .645 & .645 & $-.001$ & [.622, .695] \\
Source & .618 & .632 & .632 & $.000$  & [.613, .674] \\
High   & .904 & .745 & .655 & $-.089$ & [.641, .679] \\
\bottomrule
\end{tabular}
\par\smallskip
\textit{Qwen3.5-MoE-35B}\par\smallskip
\begin{tabular}{@{}lrrrrl@{}}
\toprule
Setting & $a$ & Pred. & Phys. & Resid. & Phys. 95\% CI \\
\midrule
Low    & .512 & .837 & .865 & $+.028$ & [.853, .913] \\
Source & .611 & .820 & .820 & $.000$  & [.808, .862] \\
High   & .901 & .921 & .718 & $-.203$ & [.709, .741] \\
\bottomrule
\end{tabular}

\end{minipage}
\end{center}

\FloatBarrier
\end{document}